\documentclass[conference]{IEEEtran}

\usepackage{tikz}
\usepackage{amsmath}
\usepackage[ruled]{algorithm}
\usepackage{algpseudocode}
\usepackage{subfigure}
\usepackage{array}
\usepackage{soul}
\usepackage{adjustbox}
\usepackage{balance}

\usepackage{amssymb}
\usepackage{pifont}

\DeclareMathOperator{\score}{score}
\DeclareMathOperator{\pred}{pred}
\DeclareMathOperator{\aand}{and}

\usepackage{fancybox}

\algdef{SE}[SUBALG]{Indent}{EndIndent}{}{\algorithmicend\ }%
\algtext*{Indent}
\algtext*{EndIndent}

\newcommand{\ie}{{\em i.e., }}
\newcommand{\eg}{{\em e.g., }}
\newcommand{\Null}{\textit{Null}}
\newcolumntype{P}[1]{>{\centering\arraybackslash}p{#1}}
\newcolumntype{V}{ >{\centering\arraybackslash} m{1.4cm} }
\newcommand{\myverb}{\fontsize{10}{48}\usefont{OT1}{lmtt}{b}{n}\noindent }
\newcolumntype{L}[1]{>{\raggedright\let\newline\\\arraybackslash\hspace{0pt}}m{#1}}
\newcolumntype{C}[1]{>{\centering\let\newline\\\arraybackslash\hspace{0pt}}m{#1}}
\newcolumntype{R}[1]{>{\raggedleft\let\newline\\\arraybackslash\hspace{0pt}}m{#1}}

\begin{document}
%
\title{AdIoTack: Quantifying and Refining Resilience of Decision Tree Ensemble Inference Models against Adversarial Volumetric Attacks on IoT Networks}

\author{
	\IEEEauthorblockN{Arman Pashamokhtari, Gustavo Batista, and Hassan Habibi Gharakheili}
	\IEEEauthorblockA{UNSW Sydney, Australia
	\\Emails: \{a.pashamokhtari,g.batista,h.habibi\}@unsw.edu.au
}}	

\maketitle

\begin{abstract}

Machine Learning-based techniques have shown success in cyber intelligence. However, they are increasingly becoming targets of sophisticated data-driven adversarial attacks resulting in misprediction, eroding their ability to detect threats on network devices. In this paper, we present \textit{AdIoTack}\footnote{Funding for this project was provided by CyAmast Pty Ltd.}, a system that highlights vulnerabilities of decision trees against adversarial attacks, helping cybersecurity teams quantify and refine the resilience of their trained models for monitoring and protecting Internet-of-Things (IoT) networks. In order to assess the model for the worst-case scenario, AdIoTack performs white-box adversarial learning to launch successful volumetric attacks that decision tree ensemble network behavioral models cannot flag. Our \textit{first} contribution is to develop a white-box algorithm that takes a trained decision tree ensemble model and the profile of an intended network-based attack (\eg TCP/UDP reflection) on a victim class as inputs. It then automatically generates recipes that specify certain packets on top of the indented attack packets (less than 15\% overhead) that together can bypass the inference model unnoticed. We ensure that the generated attack instances are feasible for launching on Internet Protocol (IP) networks and effective in their volumetric impact. Our \textit{second} contribution develops a method to monitor the network behavior of connected devices actively, inject adversarial traffic (when feasible) on behalf of a victim IoT device, and successfully launch the intended attack. Our \textit{third} contribution prototypes AdIoTack and validates its efficacy on a testbed consisting of a handful of real IoT devices monitored by a trained inference model. We demonstrate how the model detects all non-adversarial volumetric attacks on IoT devices while missing many adversarial ones. The \textit{fourth} contribution develops systematic methods for applying patches to trained decision tree ensemble models, improving their resilience against adversarial volumetric attacks. We demonstrate how our refined model detects 92\% of adversarial volumetric attacks.
\end{abstract}
	
\section{Introduction}

IoT adoption is on the rise in both consumer and business mainstreams. Still, more than half of the connected IoT devices are found vulnerable \cite{Unit42IoT2020} to a wide range of sophisticated cyber threats like botnets, malware, phishing, or DDoS attacks. 
According to a report recently published by Nokia \cite{Nokia2020}, IoTs saw a 100\% increase in infections in 2020 over the previous year. These vulnerabilities at scale can lead to significant disruption of critical enterprise operations \cite{FloridaWater2021,sierra2021,CampusVending2017} or exfiltration of sensitive data \cite{Exfilt2020}. The lack of a real-time and detailed inventory of connected IoT assets, deployed in large numbers, leads enterprises to operate their network partially blind \cite{ArunanTMC,TDSC2020Ayyoob}. This leaves vulnerable devices unmonitored and hence exposes their organization to grave risks \cite{PARVP2021}. 

While manufacturers are assumed (expected) to embed appropriate safeguards in the devices for securing them, many IoT devices have shown \cite{Unit42IoT2020, Nokia2020} to be unprotected and can be compromised with little effort from attackers. This paper advocates ``network-level'' security measures, instead of ``device-level'' \cite{Verimatrix}. We note that embedded security implementation can be highly variable across various IoT devices depending on manufacturers, device capabilities, and mode of operation. Therefore, the network-level monitoring approach comes with a number of advantages including: (a) it can be applied to a range of heterogeneous IoT devices; (b) it can be implemented, operated, and upgraded in the cloud by network operators with no dependency to device manufacturers; and (c) it can augment any device-level security implemented by the manufacturer, providing an extra layer of protection.

Given the speed and complexity of modern cyber threats, network security teams are increasingly applying machine learning (ML) techniques to network traffic (packets and/or flows) of IoT devices to model their network behavior \cite{cyberreport}. ML-based models \cite{TNSM2020} are used on the network (running on general computers fed by traffic features) to automatically classify assets from identifiable patterns in their network activity and detect anomalous behaviors \cite{Hamza2019,TDSC2020Ayyoob,Arunan2020}, indicative of compromise, firmware upgrade, or emerging novel attacks. 
Learning-based methods offer the ability to respond to situations not explicitly encountered before, replacing processes that would have required formidable manual analysis by human experts.

In the context of IoT cybersecurity, well-trained ML models have proven to effectively capture the intended behavior of IoT devices on a per-type basis. These models can flag deviations in the volume and/or frequency of network activity without being impacted by limited patterns of certain known attacks~\cite{diot,Arunan2020,Hamza2019}. This approach is successful primarily because IoT devices display a finite set of activities (with reasonably identifiable patterns) on the network during their regular operation. These behavioral characteristics present an opportunity to train models with purely benign instances obtained from IoT network traffic. The trained models would have the ability to distinguish clearly the ``bounded'' set of benign behavior from an ``unbounded'' set of malicious (anomalous/unintended) behavior resulted from of a network-based cyber attack. This gives a significant advantage to ML-based methods against traditional signature-based ones to infer from IoT network traffic.

Traditionally, the ``security of ML-based models'' has not been the main objective for designing algorithms and developing inference systems in various domains, especially in cybersecurity. Therefore, potential adversaries with certain incentives aim to subvert the ML model either during training or operation which is known as adversarial attack. Attackers may poison the training instances to influence the resulted model. They may attempt to carefully manipulate network traffic at run-time to flip predictions, yielding a poor performance of the inference model in distinguishing the malicious instances.

Several countermeasures have been proposed, under the umbrella of Adversarial Machine Learning (AML), with practical successes in the area of machine vision and image recognition~\cite{xie2019, wang2019, dhillon2018}. However, in the context of cybersecurity, it is in its nascent stage of development to the best of our knowledge. A recent report by McAfee~\cite{McAfee2019} highlights a few malware families that have bypassed machine learning engines in 2018. It predicts how cyber-criminals will be increasingly employing artificial intelligence techniques to evade detection.  

The primary objective of adversarial cyber-attackers is to find loopholes in ML-based network security models like traffic classifiers, anomaly detectors, or intrusion detection systems. To fortify inference models against these targeted attacks, cyber-security teams need to quantify the resilience of their trained model and refine any loopholes. This paper focuses on developing techniques for launching volumetric attacks (known as adversarial evasion attack) on IoT devices without being noticed by a decision tree ensemble inference model that continuously monitors the network traffic. To make these ideas more concrete, let us consider a scenario whereby an attacker aims to launch a TCP SYN reflection attack (common in large-scale DDoS attacks sourced from IoT devices \cite{synDDoS}) with 1000 attack packets. 

The attacker sends 1000 TCP SYN packets with spoofed source IP address to a target IoT device which replies (reflects) them to the victim destination (the source entity specified in the SYN packets) with 1000 TCP SYN-ACK packets. IoT devices typically generate a small amount of traffic volume on their limited set of TCP/UDP flows; thus, reflection attacks with high rates can be detected by inference models relatively easily. That said, attackers with sufficient \textit{prior knowledge} would have the ability to launch their intended attack (without being detected) by precisely generating and injecting some adversarial network traffic (say, 50 NTP packets along with 5 DNS packets and an SSDP packet) in addition to the intended attack packets (1000 TCP SYN packets). To humans, this additional traffic seems to be irrelevant and random; however, this well-crafted ``\textit{recipe}'' subverts the model's internal decision-making, leading it to accept this network traffic instance as benign. Finding these attack recipes would depend on the inference model's traffic features and detection algorithm, which requires a well-developed adversarial learning algorithm to generate them precisely.
It is important to note that this paper uses the term ``attack'' in two different contexts. First is the network-based volumetric attack (\eg TCP SYN reflection) on IoT devices, whereby target IoT devices reflect the incoming attack traffic towards an external victim. Second is the adversarial machine learning attack which is a technique that utilizes prior knowledge to subvert the ML-based inference model, and hence the volumetric attacks can go undetected.

In the literature, researchers have studied adversarial attacks on IoT networks~\cite{Ferdowsi2019, Vahakainu2020, Ibitoye2019, Sagduyu2019, Caminero2019}. Prior works primarily focus on developing adversarial models that can learn how to bypass neural network-based attack detection models. In contrast, the literature has not devoted much attention to decision tree-based models. Unlike neural networks, decision tree-based models are not differentiable; thus, well-known approaches like~\cite{Goodfellow2014, Papernot2016, Croce2020} are not compatible with their structure. Additionally, given the dynamics of network traffic and some constraints on Internet Protocol (IP) packets, adversarial attacks in the context of network security (unlike image processing) become relatively more challenging. Another gap in the current literature is that adversarial attacks, to the best of our knowledge, have never been executed in a real network protected by an ML-based inference model. Lastly, no prior systematic attempt has been made to improve the resilience of decision trees against these sophisticated attacks without manipulating the training process which requires re-training the model.

This paper presents AdIoTack, a systematic approach for learning and launching adversarial attacks on IoT networks protected by decision tree ensemble models. We make four main contributions: \textbf{(1)} We develop a novel adversarial learning algorithm named offline learning that automatically generates adversarial attack ``recipes'' given an intended volumetric attack on a target class, subverting a trained decision tree ensemble model (\S\ref{sec:adiotack}). We consider two representative network-based volumetric attacks, namely TCP SYN and SSDP reflection, which are widely used on IoT devices for launching DoS/DDoS attacks \cite{IBMsec,synDDoS,Wisec2017,NETSCOUT}. Adversarial recipes specify certain overhead packets ($2$\% for TCP SYN reflection and $14$\% for SSDP reflection) to be injected for an intended volumetric attack to go unnoticed; \textbf{(2)} We develop an online execution method that launches the adversarial instances (learned from offline learning) on a real IoT device network monitored by a decision tree ensemble inference model (\S\ref{sec:online}); \textbf{(3)} We demonstrate the performance of AdIoTack by applying it to our testbed comprising of nine consumer IoT devices. 
We show our online attack execution has at least a $95$\% chance of successfully launching an adversarial attack when the attacker has the prior knowledge of the inference model cycles (\S\ref{sec:eval}); and, \textbf{(4)} We develop a method for patching decision trees (without a need for re-training), making them more robust against adversarial volumetric attacks while maintaining the inference accuracy for benign traffic (\S\ref{sec:patching}). 
Our refined model can detect all adversarial SSDP reflection attacks with low (less than 200 attack packets per minute), medium (between 200 and 700 attack packets per minute), and high (greater than 700 attack packets per minute) impact rates. For adversarial SYN reflection attacks, the refined model detects $63\%$ of low, $75\%$ of medium, and $100\%$ of high impact attacks.



\section{Related Work}
\textbf{Cyber Intelligence for IoT Infrastructure:}
The cyber-security of IoT networks differs from that of non-IoT or traditional IT networks where general-purpose computers, smartphones, or tablets display an unbounded range of network behaviors reflecting their users' online activity. IoT devices, in contrast, have a limited range of activities (with slight  variations but predictable) in the traffic pattern. While modeling benign behavior of non-IoTs is complicated (or even impossible), IoT devices' intended activity can be formally defined and enforced on the network. A ``whitelist'' of IoT behaviors, specified in the form of MUD (Manufacturer Usage Description) profiles, has been employed to detect anomalies \cite{TDSC2020Ayyoob}. Researchers have employed various ML techniques to learn from the benign behavior of IoT devices to monitor their health and detect malicious incidents \cite{diot, Arunan2020, Hamza2019,LCN2021}. The use of pure benign instances from IoT network traffic in training inference models, without the inclusion of known attack (malicious) instances, enables them to become more robust against unseen and morphing cyber-attacks.

\vspace{-1mm}	
\textbf{Adversarial Attacks:}
Adversarial attacks were first studied in the context of image recognition, where the objective is to create an adversarial copy of a benign image that is indistinguishable to human eyes. However, the image classifier mispredicts the adversarial copy. Adversarial learning in cybersecurity differs from that in the image recognition domain in (1) in image recognition, the similarity of the adversarial instance and the benign instance, expected to look identical to human eyes, is the primary constraint of the problem. However, this constraint is relaxed in cybersecurity and traffic inference problems; (2) there is no requirement for the impact or intensity of intended attacks in image recognition problems, \ie any adversarial instances (with any level of deviations from the benign instance) are accepted as long as they subvert the model; (3) Practicality of adversarial attacks (\ie executing them in a real environment) is of concerns in cybersecurity since every adversarial instance (a set of numerical values) may not be necessarily realized as network packets.

Adversarial attack on differentiable models like neural networks has been widely studied both in image recognition and cybersecurity research problems, while non-differential models like decision tree-based models remain relatively under-explored. 
Also, adversarial learning techniques (computing gradients of the model's loss function) for differentiable models cannot be readily applied to non-differentiable models.   
Work in \cite{Ding2019} developed a method that replaces last few layers of a neural network model with a Random Forest, hide the gradients of the neural network and perhaps protecting it against adversarial attacks.

These prior works have studied evasion attacks on decision tree-based models \cite{chen2019robust, Kantchelian2016, zhang2020}. Authors in \cite{chen2019robust} has shown vulnerability of Gradient Boosting and Random Forest models against adversarial evasion attacks using methods of \cite{cheng2018, Kantchelian2016}. They also developed a robustness technique to avoid adversarial attacks by bringing adversarial instances to the training phase to change the ensemble model with this purpose that the model still has high accuracy for attack inputs. Work in \cite{Kantchelian2016} developed a mixed-integer linear programming (MIPS) technique for solving the adversarial attack optimization problem on tree ensembles. All existing works, applied their methods on public dataset like MNIST \cite{MNIST}; in contrast, to the best of our knowledge, our paper is the first in developing techniques for adversarial attacks against tree ensembles in the context of IoT cybersecurity.

\vspace{-1mm}
\textbf{Adversarial Attack Approaches in IoT:}
Existing works in IoT cybersecurity primarily adopt techniques from the image recognition domain with little adaptation and contextualization. Hence, some of the fundamental challenges and differences (discussed above) remain unaddressed.
Authors of~\cite{Caminero2019} employ reinforcement learning (RL) to generate adversarial instances against their multi-class inference model (four classes of cyberattacks plus a class of benign traffic). In that work, the adversary iteratively adds random noises to samples of an attack dataset and presents them to the inference model until they get classified as benign. However, randomly generated numerical instances do not necessarily represent realistic cyberattacks.
Authors of \cite{Abusnaina2019} studied the impact of adversarial attacks on models that detect malware-infected IoT applications. Their neural network-based model infers from graph-like features of the binary file of applications. Work in~\cite{Ferdowsi2019} trained a set of binary-class (malicious versus benign) inference models, each per unit of IoT devices. The authors used GAN (Generative Adversarial Networks), a well-known adversarial learning technique for neural networks (initially designed for image classification context). The primary objective is to improve the resilience of their model during training by including adversarial instances in the training set.
Another work~\cite{Ibitoye2019} focused on neural networks for detecting network-based attacks. Their detection is done by a multi-class inference model (with four classes of cyberattacks and one class of benign traffic). Their objective was to manipulate numerical instances of either attack or benign for which the model makes incorrect predictions. The authors directly borrowed techniques from image recognition and did not demonstrate the feasibility of subverting the model on real network. Importantly, they do not attempt to refine the vulnerable inference model.





Our work distinguishes from the prior adversarial attack studies on decision tree-based models by (a) inter-dependency among many features in this paper versus tweaking only one feature in prior works; (2) considering the impact of the volumetric network attacks and their feasibility on real networks; (3) existing works did not explore the sequence order of decision trees to find adversarial instances. We, instead, analyze how the order our approach processes the decision trees in a forest model can affect the richness of adversarial instances; (4) our robustness technique is done post-training which does not require re-training the model and is not a function of a set of known adversarial instances.

Prior works in IoT cybersecurity were directly adopted from those in image recognition with their primary focus on neural network models. In this paper, we demonstrate our method of generating and launching adversarial attacks on a live network of IoT devices. To the best of our knowledge, no prior work has focused on decision tree-based models in this context while considering the practical challenges; and this is the first time that an adversarial attack is executed on a real network of IoT devices. Also, our patching techniques systematically refine trained decision trees to detect adversarial volumetric attacks on data networks whereas previous works \cite{chen2019robust, calzavara2020treant} targeted other application domains like image recognition.
\section{AdIoTack: System Architecture and Adversarial Learning}\label{sec:adiotack}
This section describes our AdIoTack system, including major functional decisions and system components (\S\ref{sec:arc}), offline learning (\S\ref{sec:learning}), the core algorithms (\S\ref{sec:algorithms}) and consistency verification (\S\ref{sec:vercon}).

\subsection{Threat Model: Functional Decisions and \\System Components}\label{sec:arc}
{\color{black} The main objective of AdIoTack is to help cyber-security teams quantify the resilience of their decision tree-based models against adversarial volumetric attacks that target \textit{integrity} of the inference models.}
Given a well-trained traffic inference model that is able to detect ``non-adversarial'' attacks relatively easily, AdIoTack generates adversarial instances that the model mispredicts as benign. Note that {\color{black} ML models may be tested for other aspects like their \textit{availability}, \ie whether they remain operational while being flooded by malicious requests} \cite{Barreno2010}, which is beyond the scope of this paper. In terms of influence, adversarial attacks may be either \textit{causative} or \textit{exploratory} \cite{Barreno2010}. In causative attacks, training data instances are poisoned in order to manipulate the inference logic. In exploratory attacks which happens post-training, well-crafted malicious traffic instances that resemble benign instances are used to bypass the model while launching an attack. Our focus in AdIoTack is on exploratory attacks. 

{\color{black} One can quantify the resilience of an ML model in three different scenarios: (a)  \textit{White-box} is the worst-case scenario when the knowledge of the model's structure, algorithm, and features is used to evaluate the model against adversarial attacks -- testing and improving the model's performance by a white-box approach would result in the best resilience \cite{CW2017}; (b) \textit{Grey-box}, when the model is evaluated with limited knowledge of the model (say, partial access to the training data); (c) \textit{Black-box}, when the model is attacked without any prior knowledge, and it can only be probed \eg through API calls. This paper aims to help cyber-security teams evaluate and refine their decision tree-based models (self-assessment). Therefore, we develop AdIoTack in a white-box setup to maximise effectiveness in a worst-case scenario. 
}

\begin{figure}[t!]
	\centering
	\includegraphics[width=0.93\linewidth]{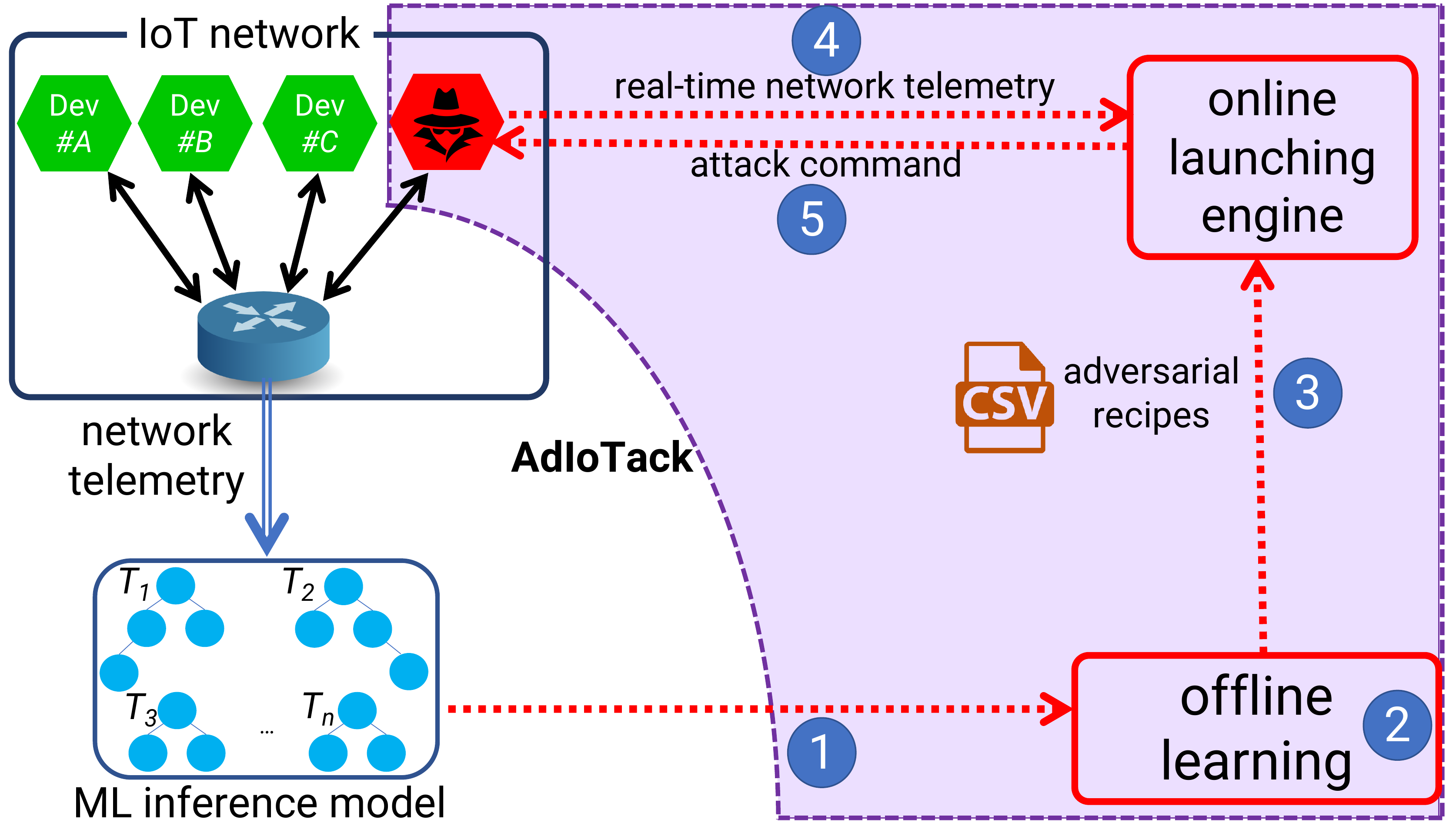}
	\vspace{-2mm}
	\caption{System architecture of AdIoTack.}
	\label{fig:arch}
	\vspace{-3mm}
\end{figure}

Fig.~\ref{fig:arch} shows the architecture of our AdIoTack system. On the top left, we see a network of IoT devices (green boxes) whose traffic is monitored periodically (\eg every one minute) by a decision tree ensemble model (\ie trees $T_1$ to $T_n$). The network router extracts the required telemetry (\eg packet/byte count of different traffic flows) from the network traffic of each IoT device and passes it to the ML inference model (bottom left). The model continuously infers the most probable class of the connected devices from the received telemetry. We use the classification score of the ML model to compute a device-specific threshold to distinguish benign behavior from malicious. That way, if the model yields a score below the given threshold, it is an indication for malicious behavior (details discussed in \S\ref{sec:offline_learning}). For AdIoTack (shown by shaded region), there are two essential phases, namely: (a) learning and (b) execution (launching). 

In the learning phase (steps \ovalbox{1} and \ovalbox{2} in Fig.~\ref{fig:arch}), the offline learning module uses the inference model (note the white-box scenario) to learn blind spots of the model to generate adversarial recipes (step \ovalbox{3}). Each adversarial recipe consists of a number of conditions over features of the inference model (\eg $ 10 < f_1$, $f_2 \le 50$, and $20 < f_3 < 40$); thus, each recipe can generate several adversarial instances that conform to the recipe's conditions. For example, an adversarial instance for the mentioned recipe could be $f_1 = 15$, $f_2 = 30$, and $f_3 = 25$. For execution, network telemetry (step \ovalbox{4}) similar to inputs of the inference model must be collected, so that an ``appropriate'' adversarial recipe can be selected based on the current state of the network. To obtain real-time telemetry and execute the intended adversarial attack, network traffic of IoT devices is snooped. As a part of our threat model, we assume a ``malicious agent'' like an infected smartphone or a computer is already inside the local network (the red box inside the IoT network on the top left) for this purpose.


To obtain network telemetry in real-time, the malicious agent can passively (\eg sniffing) or actively (\eg man-in-the-middle via ARP poisoning) monitor the behavior of victim IoT devices. The passive approach is stealthier to perform but could be practically challenging in some cases, like when the malicious agent and its victims are on different physical access mediums (wireless versus wired). For AdIoTack, we employ the active mode for adversarial network monitoring. Also, our method only analyzes packet metadata (headers), so it has no issue with encrypted payloads.

In the next step, the online engine starts a search for feasible adversarial recipes, given the current state of the network. From those candidate recipes, the online engine selects the closest one to the current state of the network to minimize the amount of adversarial packets (overhead) to be injected (in real-time to the network traffic) on behalf of the victim. Eventually, the intended volumetric attack is launched (step \ovalbox{5}) by judiciously crafting network traffic matching the chosen recipe. Note that the local victim will reflect/amplify the attack traffic onto an ultimate victim on the Internet.
It is important to note that the adversarial packets (overhead traffic) are injected on behalf of local victims for their reflected/amplified traffic to go undetected.


\subsection{Offline Adversarial Learning}\label{sec:learning}
\label{sec:offline_learning}
Offline adversarial learning is a process of generating adversarial recipes which: (a) result in desired malicious impact, and (b) can go undetected \ie the traffic inference model raises no anomaly flag. The results of offline adversarial learning show to what extent the given model is vulnerable to adversarial volumetric attacks.

We employ ML models' \textit{classification score} as a measure for detecting behavioral changes. We define a classification score threshold for each class (device type) according to the training results. At run-time, any instance with a classification score below the given threshold is flagged as an anomaly. We can define the adversarial learning problem formally as below:

\begin{gather*}
\label{eq:adv}
\forall{d \in \textbf{D}},~\{x~|~\pred(x) = d ~\aand~ \score(x) \ge T_d\} \\
\textrm{s.t.}~~~\forall{f \in \textbf{F}},~x_f \ge f_{min}
\end{gather*}

This equation defines a set of adversarial instances $\textbf{X}$ for each IoT device type $d \in \textbf{D}$. The output of the inference model for each adversarial instance $x \in \textbf{X}$, $\pred(x)$, is $d$ and the classification score, $\score(x)$, is greater than or equal to the classification threshold of the corresponding class, $T_d$. Therefore, every adversarial instance $x$ can bypass the inference model's detection as it satisfies the minimal classification score. Also, we define a set of constraints to guarantee that $\textbf{X}$ fulfills the requirements of the desired attack. As the scope of this paper is volumetric attacks, we define these constraints as a set of lower bound conditions for each feature $f$ that belongs to the target feature set $\textbf{F}$. The set $\textbf{F}$ contains the relevant features that would be affected by the desired attack, \eg TCP packet and byte count features in TCP SYN reflection attack.


\begin{figure}[t!]
	\centering
	\includegraphics[width=\linewidth]{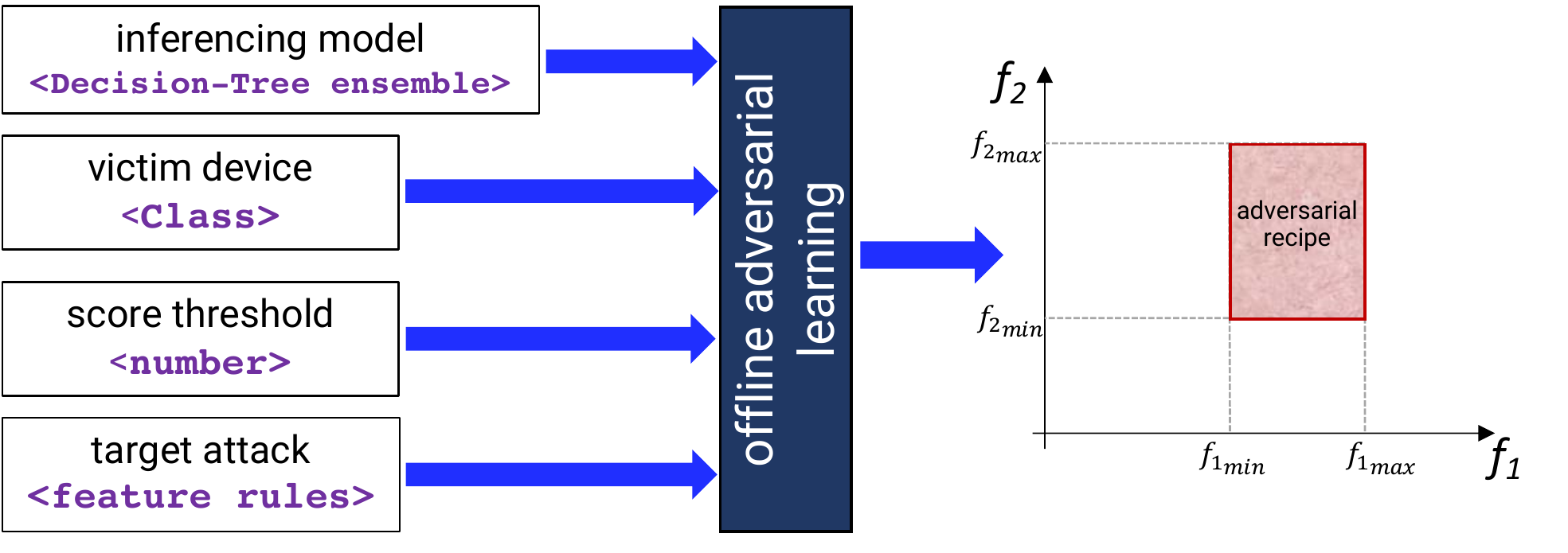}
	\vspace{-4mm}
	\caption{Inputs and outputs of our adversarial learning process.}
	\label{fig:adv_process}
	\vspace{-2mm}
\end{figure}

\begin{figure*}[t!]
	\begin{center}
		\includegraphics[width=0.83\linewidth]{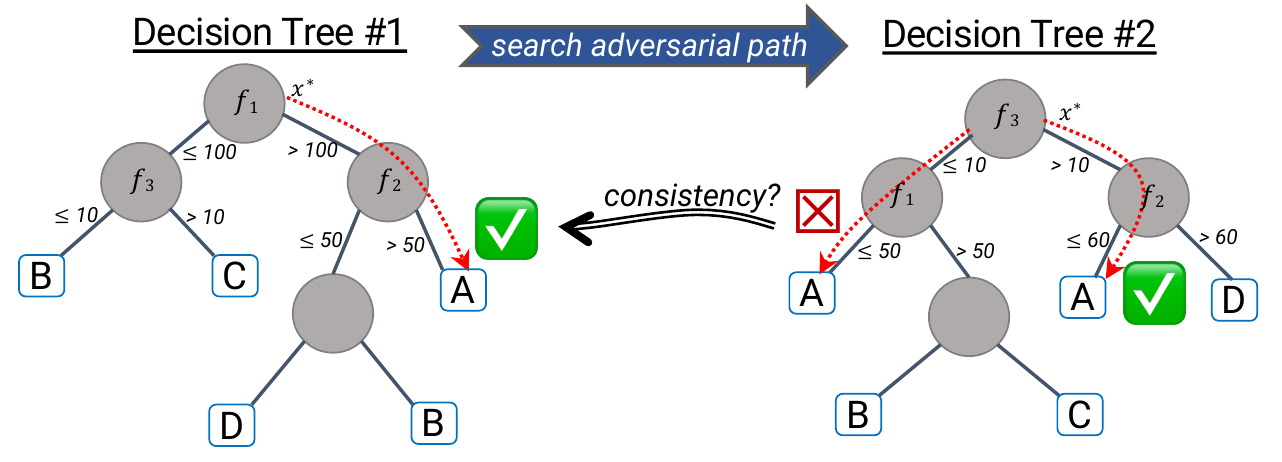}
		\vspace{-4mm}
		\caption{Searching adversarial paths in two decision trees: The adversary starts traversing decision tree \#1 with no prior rule -- a path to target class ``{\myverb{A}}'' is found ($\checkmark$). Moving to decision tree \#2, the first path becomes inconsistent with rules of the chosen path from decision tree \#1 ($\times$), and thus an alternative path is sought and found ($\checkmark$).}
		\label{fig:RF_adv_path}
		\vspace{-6mm}
	\end{center}
\end{figure*}

Fig.~\ref{fig:adv_process} illustrates inputs and outputs of our offline adversarial learning process. For illustration purpose we only show two features, $f_1$ and $f_2$. Suppose we want to test whether the model is vulnerable to a volumetric attack on a victim device (\eg Amazon Echo), where the attack offers more than 1000 packets over $f_1$. The desired attack impact is captured by a set of conditions for the target feature(s) (in our example, $f_1 > 1000$). The offline adversarial learning process then searches over a given tree ensemble model in order to determine recipes that the ensemble model predicts them as the class of the victim device with a classification score of greater than or equal to the given threshold. Adversarial recipes can be seen as a region in an $N$ dimension space, being $N$ the number of features required by the inference model. Conditions on target features ($f_1$ in our example) specified by the adversarial recipes must be consistent with those conditions given as input for the intended attack (\ie $f_1 > 1000$). In other words, any point in an adversarial recipe region is an adversarial instance that satisfies the conditions of the intended attack.

We can use different search methods to find $\textbf{X}$ depending on the underlying inference model. For models like neural networks, logistic regression, and SVM, this optimization problem can be solved using gradient-based methods such as gradient descent, fast gradient sign \cite{Goodfellow2014}, Jacobian saliency map \cite{Papernot2016}, or auto projected gradient descent \cite{Croce2020}. While differentiable models like neural networks have been widely applied to image processing tasks, non-differentiable models like decision trees have received less attention. In cybersecurity applications, instead, decision tree-based models seem more attractive with some interesting properties: (i) faster training of $\mathcal{O}(n\log n)$; (ii) better handling of imbalanced datasets \cite{Khoshgoftaar} without the need for data normalization or scaling prior to training; (iii) Random Forest algorithm is reasonably resilient to the overfitting problem \cite{breiman2001}; (iv) Decision tree-based models often yield good results in terms of accuracy and explainability which means their internal decision-making process is visible during testing, which is highly desirable for cyber-analysts (particularly for IoT network security \cite{Hasan2019, Doshi2018, Miettinen2017a, TNSM2020, Sivanathan2017}) in taking remedial and/or preventive actions. In contrast, competitors like neural networks do not provide insights into their internal process of inference; (v) Random Forest classifiers are more robust than $k$-nearest-neighbors ($k$-NN), decision trees, and AdaBoost \cite{Katzir2018} against adversarial evasion attacks, hence, become more difficult to evade. 


\subsection{The AdIoTack Algorithms}
\label{sec:algorithms}

AdIoTack focuses on performing adversarial attacks on a given decision tree ensemble model. The model consists of several decision trees. Weighted or unweighted voting gives the final classification, \ie the fraction of trees yielding the final output determines the prediction and classification score.


In order to explain how voting works, let us consider an example relevant to our device class inference problem. Let say, if a benign traffic instance $x$ belongs to IoT device class ``{\myverb{A}}'' (ground-truth), and say 95\% of the trees in the ensemble model classify it correctly, the final prediction would be ``{\myverb{A}}'' with the score of 0.95. Now, assume ``{\myverb{A}}'' is under a cyberattack, having a traffic instance $\hat{x}$. The model is expected to give a lower score \cite{TNSM2020} for $\hat{x}$ since because of the deviation from benign instance $x$ there will be disagreement among various trees, each inferring from a specific set of features. Certain features of $\hat{x}$ will significantly deviate from their expected normal range, leading a portion of trees to classify $\hat{x}$ as a class other than ``{\myverb{A}}'', and other features may remain relatively unchanged hence some other trees predict the class of $\hat{x}$ as ``{\myverb{A}}''. Therefore, the attack can be flagged with a good chance. In adversarial attack, we aim to find a recipe which has the same malicious impact of $\hat{x}$ in such a way that the majority of decision trees predict it as ``{\myverb{A}}'', resulting in high score and implying it is still benign.

\vspace{-1mm}
In the context of decision trees, we define an \textit{adversarial path} to be the path from the root node of a given decision tree to a leaf with the label of the victim device that we aim to launch an attack on it. Each adversarial path is associated with a set of conditions that are met across the path nodes. Fig.~\ref{fig:RF_adv_path}-\textit{left} shows how the adversarial path is found on an illustrative decision tree. Grey circles are decision nodes, each checking a certain condition (\eg $f_1 \le \tau_1$), proceeding to subsequent branch (left or right, depending on the check result). Adversarial path $x^*$ is a complete path from the root to a leaf node, landing on the target victim class ``{\myverb{A}}''. AdIoTack performs tree traversals using the pre-order method -- the root node is first visited, then recursively a pre-order traversal of the left subtree is performed, followed by a recursive pre-order traversal of the right subtree. 

\vspace{-2mm}
Note that searching the adversarial path needs to be extended to all decision trees inside the ensemble model where consistency of conditions across trees becomes a non-trivial challenge. Fig.~\ref{fig:RF_adv_path} shows an illustrative example of searching for an adversarial path across two decision trees while conditions (of paths in the two trees) need to be consistent. Assume we want to target class ``{\myverb{A}}'' with a cyberattack that requires $\{f_1 > 1000\}$ (\ie target rule). Below, we show the initial recipe to begin the attack:
\[
Recipe: f_1 > 1000 \Rightarrow A
\]

Starting from decision tree \#1 (on the left), AdIoTack finds the first path leading to a leaf with the target class ``{\myverb{A}}'' (by traversing the tree in pre-order mode). The adversarial path on decision tree \#1 consists of two conditions: $\{f_1 > 100$ and $f_2 > 50\}$, which are consistent with the initial recipe; therefore, we can add them to the recipe:
\[
Recipe: f_1 > 1000~\wedge~f_1 > 100~\wedge~f_2 > 50 \\ \Rightarrow A
\]

\noindent which can be simplified (\textit{merged}) to:
\[
Recipe: f_1 > 1000~\wedge~f_2 > 50 \Rightarrow A
\]

Moving on to decision tree \#2, the first possible path (in pre-order search) to the target class ``{\myverb{A}}'' requires $\{f_1 \le 50\}$, which is inconsistent with $\{f_1 > 1000\}$ in the current recipe -- this path cannot be taken, and hence an alternative path needs to be sought. The other path to the other leaf with class ``{\myverb{A}}'' (on decision tree \#2) results in two new conditions: $\{f_3 > 10$ and $f_2 \le 60\}$, not violating the current recipe determined by the path from decision tree \#1. The updated recipe is shown below:
\[
Recipe: f_1 > 1000~\wedge~f_3 > 10~\wedge~50 <f_2 \leq 60 \Rightarrow A
\]

To generate an adversarial recipe that drives the majority vote to the target class, AdIoTack needs to keep track of conditions and progressively augment them as new adversarial paths are found across individual decision trees. Our search procedure visits each tree in a greedy manner, \ie it skips a tree if it could not find any consistent adversarial path on it. Also, the order of the trees influences the output of the method. We address this problem by running the algorithm multiple times with different tree ordering to generate new recipes.

\begin{algorithm}[t!]
	\caption{Adversarial evasion attack on tree ensemble model.}\label{alg:adv}
	\footnotesize
	\begin{algorithmic}[1]
		\State {$\mathcal{IM} \gets$ inference model}
		\State {$TR \gets$ set of target rules}
		\State {$TD \gets$ target device class}
		\State {$T_d \gets$ classification score threshold of $TD$}
		\Function{FindRecipe}{$\mathcal{IM}$, $TR$, $TD$, $T_d$}
		\State {$R \gets (TR \Rightarrow TD)$} \Comment{Recipe}
		\For{$t~in~\mathcal{IM}.trees$}
		\State {$AdvPath \gets$ \Call{FindAdvPath}{$t.root$, $TD$, $R$, $[~]$}}
		\If{$AdvPath \ne \Null$}
		\State{$R \gets$ \Call{merge}{$R$, $AdvPath$}}
		\EndIf
		\EndFor
		\State{$x^* \leftarrow$ \Call{project}{$R$}}
		\If{\Call{$\mathcal{IM}$.pred}{$x^*$} = $TD$ \& \Call{$\mathcal{IM}$.score}{$x^*$} $\ge T_d$}
		\State \Return {$R$}
		\Else
		\State \Return \Null
		\EndIf
		\EndFunction
	\end{algorithmic}
\end{algorithm}

Having these intuitive illustrations of the adversarial attack method, let us formally develop it by Algorithms~\ref{alg:adv} and \ref{alg:adv_leaf}. Algorithm \ref{alg:adv} searches for an adversarial recipe for a given target device. This algorithm expects the inference model ($\mathcal{IM}$), a set of target rules ($TR$) that are defined based on the target cyberattack, the target class of victim IoT device ($TD$), and the classification score threshold ($T_d$) as inputs.

In essence, Algorithm~\ref{alg:adv} iterates over all trees in the ensemble model ($\mathcal{IM}$). Notice the tree order in a decision tree ensemble model is arbitrary, and different orders may give different results (new recipes or no recipe at all). In~\S\ref{sec:eval}, we evaluate how different orders can affect the number of generated recipes. For each tree, the algorithm calls the function {\myverb{FindAdvPath}} in Algorithm~\ref{alg:adv_leaf} to find an adversarial path consistent with the current recipe generated so far. If such a path is found, the corresponding adversarial path's rules are merged with the current recipe rules ({\myverb{Merge}}). The following examples show how {\myverb{Merge}} function works: 

{\myverb{Merge}}$\left (f \le \tau_1,~f \le \tau_2,~\tau_1 \le \tau_2\right ) = f \le \tau_1$, or

{\myverb{Merge}}$\left (f > \tau_1,~f > \tau_2,~\tau_1 > \tau_2 \right ) = f > \tau_1$

\begin{algorithm}[t!]
	\caption{Searching for a consistent adversarial path.}
	\footnotesize
	\label{alg:adv_leaf}
	\begin{algorithmic}[1]
		\Function{FindAdvPath}{$n$, $TD$, $R$, $P$}
		\If{$n$ is $leaf$}
		\If{$n.label = target$}
		\State {\Return $P$} \Comment{Adversarial path}
		\Else
		\State \Return \Null
		\EndIf
		\EndIf
		\State {$r \gets$ \Null}
		\If{\Call{isConsistent}{$n.cond, R$} \& \Call{isConsistent}{$n.cond, P$}}
		\State {$r \gets$ \Call{FindAdvPath} {$n.left$, $TD$, $R$, $[P \wedge n.cond]$}}
		\If{$r \ne \Null$}
		\State \Return $r$
		\EndIf
		\EndIf
		\If{\Call{isConsistent}{$\lnot n.cond, R$} \& \Call{isConsistent}{$\lnot n.cond, P$}}
		\State {$r \gets$ \Call{FindAdvPath} {$n.right$, $TD$, $R$, $[P \wedge \lnot n.cond]$}}
		\EndIf
		\State \Return $r$
		\EndFunction
	\end{algorithmic}
\end{algorithm}

\vspace{-2mm}
As there is no guarantee that {\myverb{FindAdvPath}} will find an adversarial path for every tree, we need to validate the efficacy of the final recipe (whether it can bypass the model or not). We may find adversarial paths effective on only a small subset of the decision trees, or even no tree at all. In this case, the resulted adversarial instances will become ineffective in yielding the target class and/or the score threshold. We validate the efficacy of a recipe by generating a representative adversarial instance, $x^*$, using {\myverb{Project}} function. We note that every instance of a recipe would fall under the desired adversarial paths determined by Algorithm~\ref{alg:adv}; therefore, if $x^*$ can bypass the model, any other adversarial instance generated from the given recipe can do the same. 
The instance $x^*$ is presented to the model to obtain the prediction. If it is classified as the target class with a classification score greater than or equal to the specified threshold ($T_d$), the algorithm approves and returns the recipe. 

\vspace{-1mm}
Algorithm~\ref{alg:adv_leaf} (invoked from within Algorithm~\ref{alg:adv}) searches for a consistent adversarial path ($P$) to a leaf with the target class label ($TD$) on a given tree. The algorithm is a direct application of pre-order traversal for binary trees. During a search, we check for consistency ({\myverb{IsConsistent}}) between the nodes' condition and the recipe (explained in \ref{sec:vercon}). This helps to prune the search space, reducing the average execution time. Algorithm~\ref{alg:adv_leaf} assumes binary decision trees; therefore, the nodes are described by only three fields ({\myverb{left}}, {\myverb{right}} and {\myverb{cond}}). The left subtree is associated with the condition {\myverb{cond}} and the right subtree with logical negation of it ({$\lnot$\myverb{cond}}).

\vspace{-1mm}
Note that an adversary may choose to stop searching when a certain level of majority across trees (\ie classification score) is obtained. Our algorithm, instead, aggressively aims to maximize its chance of subverting the model by (greedily) searching all possible trees. Given this greedy nature of our offline adversarial learning, its time complexity is $\mathcal{O(}\mathcal{T}.\mathcal{N}.\mathcal{F})$, proportional to the number trees ($\mathcal{T}$), maximum number of nodes in the trees ($\mathcal{N}$), and total number of features ($\mathcal{F}$). It is important to note that our search across trees progresses only in a forward direction. Hence, in some cases, it may not yield any consistent adversarial path on a decision tree (in the middle of iterating over trees). In other words, if a consistent path on a tree is not found, then that tree is skipped without going backward, seeking alternative adversarial paths on the previous trees, and updating the recipe correspondingly.

\subsection{Verifying Consistency of Conditions}\label{sec:vercon}
Our adversarial learning algorithm is general and can be applied to any decision tree ensemble model. However, certain constraints may be needed when employing the algorithm for specific domain problems. Given the primary use-case of this paper is network-based volumetric attacks, three types of consistency checks are performed before taking a new branch (left or right subtree) while searching for an adversarial path:

\textbf{Single-feature consistency:} This type of consistency is required when different conditions are imposed on a single feature across trees.
Let us consider two illustrative conditions $f\le\tau_1$ and $f>\tau_2$. They are considered consistent if $\tau_2<\tau_1$, where $f$ denotes a feature that the inference model uses. 
A single-feature consistency check is required in any problem of adversarial machine learning. However, in the context of cybersecurity, we identified two additional checks essential for launching attacks; otherwise, the offline learning algorithm may generate impractical adversarial recipes that cannot be launched during the online phase.

\begin{figure}[t!]
	\centering
	\includegraphics[width=0.95\linewidth]{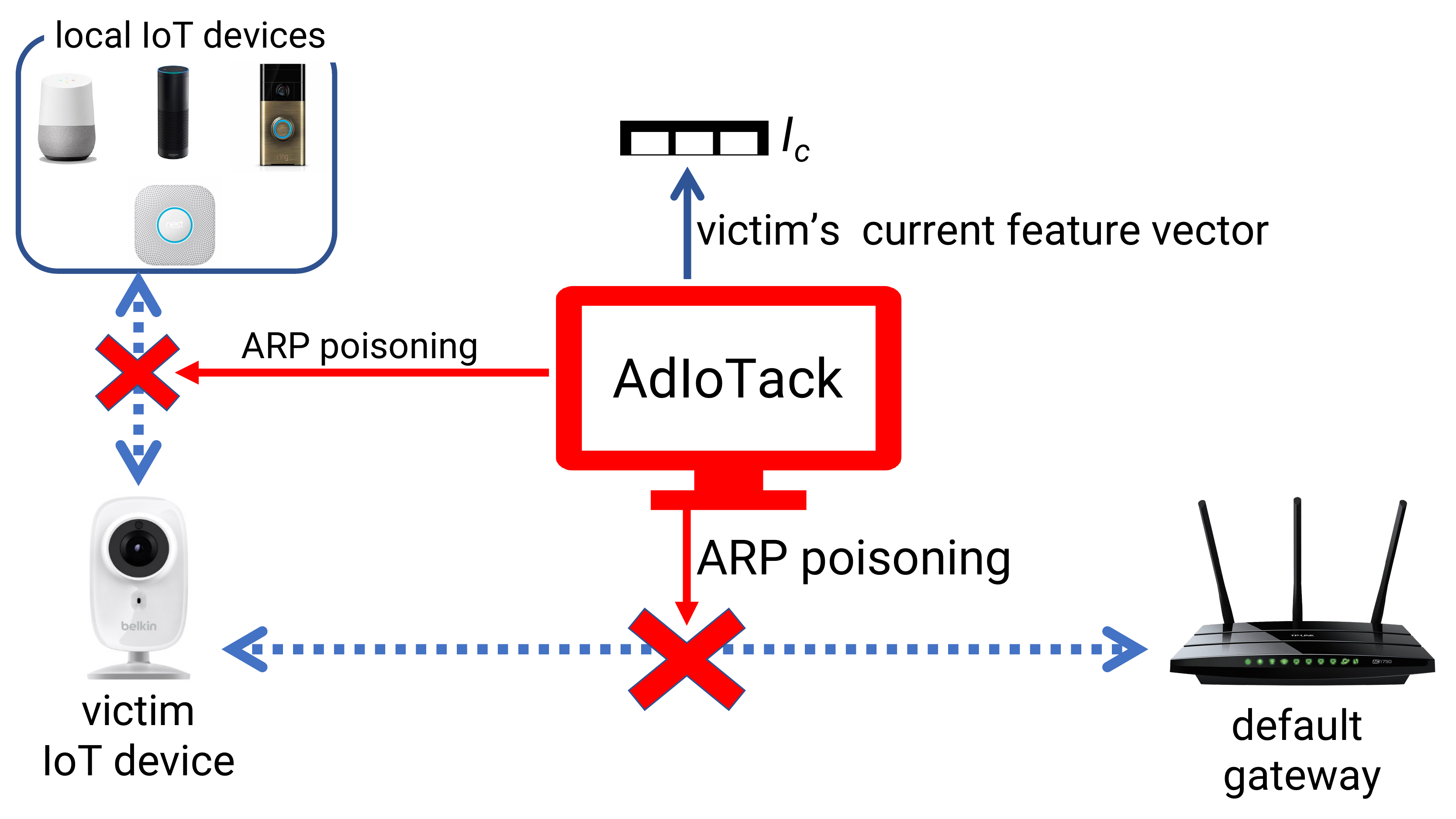}
	\vspace{-2mm}
	\caption{Local adversary (an infected machine) collects traffic features before launching an online attack.}
	\vspace{-2mm}
	\label{fig:attack_prep}
\end{figure}

\textbf{Frame size consistency:} This constraint is specific to the use-case of this paper. Frame size is measured in bytes and has a minimum and maximum length, depending on the implemented technology. For example, an Ethernet frame must be at least 64 bytes for collision detection to work and can be a maximum of 1518 bytes to avoid IP fragmentation. This expected characteristic may get violated by a condition. For example, let us consider a set of conditions on two features of incoming ($\downarrow$) DNS traffic for a given adversarial recipe:

\[
\Bigg \{
\begin{aligned}
C_1: 9<\ovalbox{$\downarrow$DNS~packet~count} \\
C_2: \ovalbox{$\downarrow$DNS~byte~count} \le 100
\end{aligned}
\]

These translate into an average size of incoming DNS frames to be less than 10 bytes, which is impossible on Ethernet networks -- sending at least ten frames even with no payload would result in 640 bytes of traffic, violating the condition $C_2$ above. 

\textbf{Boundary consistency:} Average frame size obtained by the ratio of byte count and packet count must be rounded to integer values (cannot be float values). There are certain corner cases where packet count and byte count conditions become inconsistent. The following conditions exemplify this situation:

\[
\Bigg \{
\begin{aligned}
C_3: 10<\ovalbox{$\downarrow$DNS~packet~count} \le 15 \\
C_4: 998<\ovalbox{$\downarrow$DNS~byte~count} \le 1,000
\end{aligned}
\]

Choosing the combination of \ovalbox{$\downarrow$DNS~packet~count} = 11  and \ovalbox{$\downarrow$DNS~byte~count} = 999 gives a minimum average frame size of 91 B (equally sized). At a high level, this combination would violate $C_4$ since 11 DNS incoming messages of size  91 B each will amount to 1001 B. Indeed, one may choose to solve this inequality problem by judiciously setting packet sizes (\eg nine packets of each 100 B and a packet of 99 B) -- this is beyond the scope of this paper. We only consider recipes that allow adversarial attacks of equal packet size.  

Note that in a single tree, conditions inside each path (from root to a leaf) satisfy single-feature consistency by default. Still, even conditions within a given path of a single tree may violate frame size and boundary consistencies. 

\section{Launching Adversarial Volumetric Attack}\label{sec:online}
This section describes how we use adversarial recipes to launch real volumetric attacks on an operational IoT network whose traffic is continuously monitored by a trained inference model. To clarify the method better, we explain this section with a set of the flow-based features that the model uses for inference. Inspired by \cite{TNSM2020}, we choose packet count and byte count of certain network flows: $\downarrow${\myverb{DNS}}, $\uparrow${\myverb{DNS}}, $\downarrow${\myverb{NTP}}, $\uparrow${\myverb{NTP}}, $\uparrow${\myverb{SSDP}}, $\downarrow${\myverb{LAN}}, $\downarrow${\myverb{WAN}}, and $\uparrow${\myverb{WAN}}, proven to be reflective of IoT behavior, during fixed time windows (of length one minute); where $\downarrow$ and $\uparrow$ indicate incoming and outgoing directions, respectively.

\begin{figure}[t!]
	\centering
	\includegraphics[width=0.95\linewidth]{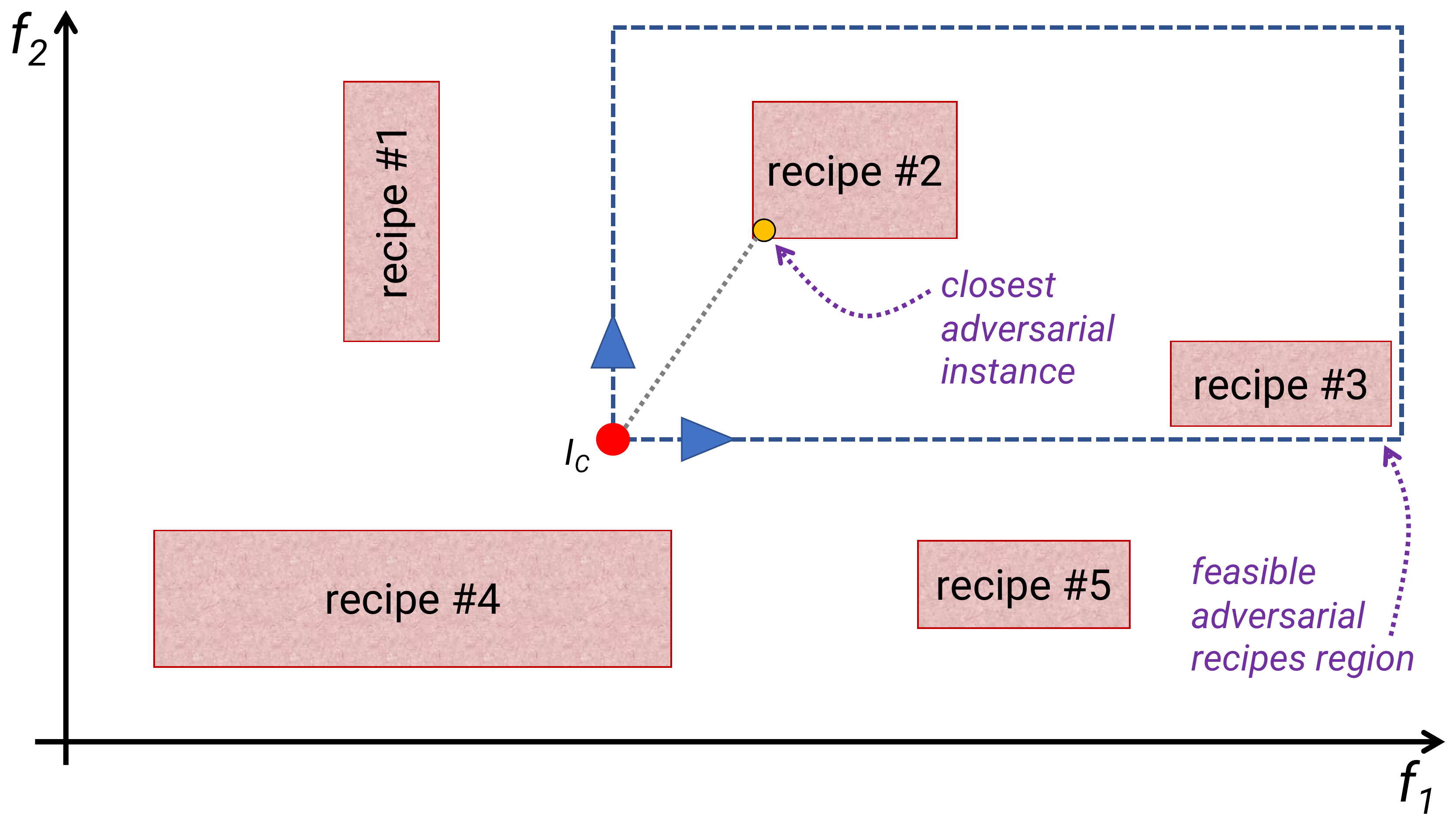}
	\vspace{-2mm}
	\caption{Finding feasible recipes' region and the closest adversarial instance given current state $I_c$. For illustration purpose, only two features are shown.}
	\vspace{-4mm}
	\label{fig:closest_advrs}
\end{figure}

\begin{table*}[htb]
	\centering
	\small
	\caption{An example of a feasible SYN reflection attack on Google Chromecast.}
	\begin{adjustbox}{max width=\textwidth} 
		\renewcommand{\arraystretch}{1.2}
		\begin{tabular}{l|C{0.5cm} C{0.8cm} C{0.5cm} C{0.8cm} C{0.5cm} C{0.5cm} C{0.5cm} C{0.5cm} C{0.5cm} C{0.8cm} C{0.9cm} C{1cm} C{0.8cm} C{1.15cm} C{0.9cm} C{0.9cm}} \hline
			
			\hline
			\multicolumn{1}{L{1cm}|}{} & 
			\rotatebox{90}{$\downarrow$ {\myverb{DNS}} pkt} &
			\rotatebox{90}{$\downarrow$ {\myverb{DNS}} byte} &
			\rotatebox{90}{$\uparrow$ {\myverb{DNS}} pkt} &
			\rotatebox{90}{$\uparrow$ {\myverb{DNS}} byte} &
			\rotatebox{90}{$\downarrow$ {\myverb{NTP}} pkt} &
			\rotatebox{90}{$\downarrow$ {\myverb{NTP}} byte} &
			\rotatebox{90}{$\uparrow$ {\myverb{NTP}} pkt} &
			\rotatebox{90}{$\uparrow$ {\myverb{NTP}} byte} &
			\rotatebox{90}{$\uparrow$ {\myverb{SSDP}} pkt} &
			\rotatebox{90}{$\uparrow$ {\myverb{SSDP}} byte} &
			\rotatebox{90}{$\downarrow$ {\myverb{LAN}} pkt} &
			\rotatebox{90}{$\downarrow$ {\myverb{LAN}} byte} &
			\rotatebox{90}{$\downarrow$ {\myverb{WAN}}  pkt} &
			\rotatebox{90}{$\downarrow$ {\myverb{WAN}}  byte} &
			\rotatebox{90}{$\uparrow$ {\myverb{WAN}}  pkt} &
			\rotatebox{90}{$\uparrow$ {\myverb{WAN}}  byte}\\ \hline
			Condition ($>$) & 2 & 119 & 2 & 194 & * & 45 & * & * & 7 & 4,353 & 94 & 9,303 & 88 & 45,568 & 10 & 5,650\\ 
			Condition ($\le$) & * & 1,666 & * & * & * & * & 1 & * & 9 & 4,553 & 148 & 13,797 & * & * & * & *\\
			Current state ($I_c$) & 0 & 0 & 0 & 0 & 0 & 0 & 0 & 0 & 0 & 0 & 0 & 0 & 46 & 125,857 & 51 & 5,752\\
			Adversarial instance & \underline{3} & \underline{120} & \underline{3} & \underline{195} & \underline{1} & \underline{46} & \underline{0} & \underline{0} & \underline{8} & \underline{4,354} & \underline{95} & \underline{9,304} & \textbf{1,000} & \textbf{74,000} & \textbf{1,000} & \textbf{74,000}\\ \hline
			
			\hline
		\end{tabular}
	\end{adjustbox}
	\label{tab:online_sample}
	\vspace{-1mm}
\end{table*}

This phase aims to project the current state (traffic feature vector) of a victim IoT device to an adversarial instance using one of the adversarial recipes obtained in the offline learning phase. A naive approach is to launch a cyberattack blindly based on a random recipe. This approach is not guaranteed to bypass the inference model given the variability in volume and frequency of traffic sent/received by the victim IoT devices during the last time window (\ie current state). It is important to note that malicious packets (pertinent to the intended network attack) get added to the benign traffic of the victim device. Hence, a more effective strategy is needed that considers the current state of the variable network.

A well-known method to track network traffic and create a feature vector for the victim device is poisoning the ARP tables (discussed in \S\ref{sec:adiotack}) of local IoT devices and default gateway. Therefore, AdIoTack can sit in the middle of these entities by forwarding traffic between them. Fig.~\ref{fig:attack_prep} illustrates the monitoring phase before launching the attack. AdIoTack maintains a record (feature vector $I_c$) of the traffic features in the current epoch. Given $I_c$, AdIoTack would find feasible adversarial recipes towards the end of each epoch. A feasible recipe has to be an upper bound to vector $I_c$ (Fig.~\ref{fig:closest_advrs} shows it in a 2D space).
If the victim displays network activities ($I_c$) exceeding a threshold ($f < \tau$) specified by a recipe, then $I_c$ cannot be projected to that recipe anymore simply because we cannot reduce the amount of traffic that already has been exchanged by the device but we can increase it.  
Among all feasible recipes, AdIoTack selects the closest adversarial instance to $I_c$, minimizing the overhead packets to be injected. This strategy requires AdIoTack to be aware of the inference model's timing cycles, \ie when it is called for prediction. In our evaluation (\S \ref{sec:eval}), we will explain how AdIoTack has the chance to execute successful attacks even without syncing with cycles of the inference model.

\begin{table}[t!]
	\centering
	\small
	\caption{Spoofed metadata for each feature.}
	\begin{adjustbox}{max width=0.48\textwidth}
		\renewcommand{\arraystretch}{1.5}
		\begin{tabular}{l|C{0.8cm}C{0.8cm}cccc} \hline
			
			\hline		
			Feature & src MAC & dst MAC & src IP & dst IP & src Port & dst Port \\ \hline
			$\downarrow$ DNS & GW & VIC & * & VIC & 53 & * \\
			$\uparrow$ DNS & VIC & GW & VIC & * & * & 53 \\ 
			$\downarrow$ NTP & GW & VIC & * & VIC & 123 & * \\ 
			$\uparrow$ NTP & VIC & GW & VIC & * & * & 123 \\ 
			$\uparrow$ SSDP & VIC & * & VIC & * & * & 1900 \\ 
			$\downarrow$ LAN & * & VIC & LAN IP & VIC & * & * \\ 
			$\downarrow$ WAN & GW & VIC & WAN IP & VIC & * & * \\
			$\uparrow$ WAN & VIC & GW & VIC & WAN IP & * & * \\ \hline
			
			\hline
		\end{tabular}
	\end{adjustbox}
	\label{tab:spoof}
	\vspace{-1mm}
\end{table}

In this paper, we use linear search to find feasible recipes. Though we do not encounter a challenge in terms of the time complexity of the linear search in our evaluation, there might be situations where a simple linear search becomes inefficient. 
To achieve certain response time, one may attempt to optimize the search process and/or prune the generated recipes. Both of these objectives are beyond the scope of this paper.


We use an example in Table~\ref{tab:online_sample} to better clarify our method for executing a volumetric TCP SYN reflection attack on Google Chromecast. The top two rows indicate lower-bound and upper-bound feature values specified by the adversarial recipe for launching a successful attack. Note that the symbol ``*'' indicates unbounded features (no lower/upper limits). The third row shows the current feature vector of Google Chromecast over the last 60 seconds. The bottom row shows the adversarial instance comprising the intended volumetric attack traffic (bold cells) reflected by Google Chromecast and overhead packets (underlined cells) injected by AdIoTack.

To successfully launch the adversarial volumetric attack, AdIoTack needs to inject certain spoofed packets (the overhead traffic specified by the closest recipe) on behalf of the victim, so that the inference model does not detect any deviation in the behavior of the victim IoT device.
Table~\ref{tab:spoof} summarizes certain packet metadata fields to be modified along with their spoofed value for a given set of traffic features in the specific model we studied. GW and VIC stand for gateway and victim, respectively. WAN IP and LAN IP respectively refer to external (public) and internal (private) IP address. Also, ``*'' highlights a wildcard value for the respective field.

\vspace{-1mm}
\textbf{Practical Challenges of Network-Based Attacks:} We discuss a few practical challenges to overcome for successful execution of adversarial attack instances in what follows:
(i) in order to intercept bidirectional traffic exchanged between the victim and its local network, AdIoTack needs to employ Gratuitous ARP replies to send a broadcast spoofed ARP reply (on behalf of the victim device) to all local devices. That way, all local traffic destined to the victim device is forwarded to AdIoTack; and, 
(ii) there might be cases whereby AdIoTack and victim may not share the same network interface (\eg one is wired and the other one is wireless). In that case, AdIoTack will have to send the spoofed traffic via the default gateway, which will disrupt the MAC-learning tables (wireless vs. wired) of the default gateway's interfaces due to the spoofed source MAC address in those packets (Table~\ref{tab:spoof}) -- this can lead to mis-forwarding of other packets to/from the victim (wireless LAN instead of wired LAN). Disrupting the default gateway's MAC tables results in dropping future packets with the same MAC address, this time as the destination field. To avoid this, AdIoTack sends a specific {\myverb{ICMP}} packet with destination broadcast MAC address ({\myverb{ff:ff:ff:ff:ff:ff}}) and victim's IP address. 
The victim's reply will revert the MAC tables to their correct state.
\section{Evaluation Results}
\label{sec:eval}
This section evaluates the performance of AdIoTack in its two phases, namely offline adversarial learning, and online adversarial execution. We begin by training three well-known decision tree ensemble models namely Random Forest, Gradient Boosting, and AdaBoost classifiers.




\begin{table}[t!]
	\centering
	\small
	\caption{Three ensemble models performance on the testing dataset.}
	\begin{tabular}{lcc} \hline
		\hline
		\textbf{Model} & \textbf{Accuracy} & \textbf{False positive} \\ \hline 
		Random Forest & 96\% & 9\% \\ 
		Gradient Boosting & 91\% & 0.1\% \\ 
		AdaBoost & 99\% & 5\%\\ \hline
		\hline
	\end{tabular}\label{tab:modelsPerformance}
	\vspace{-1mm}
\end{table}

\begin{figure}[t!]
	\centering
	\includegraphics[width=0.93\linewidth]{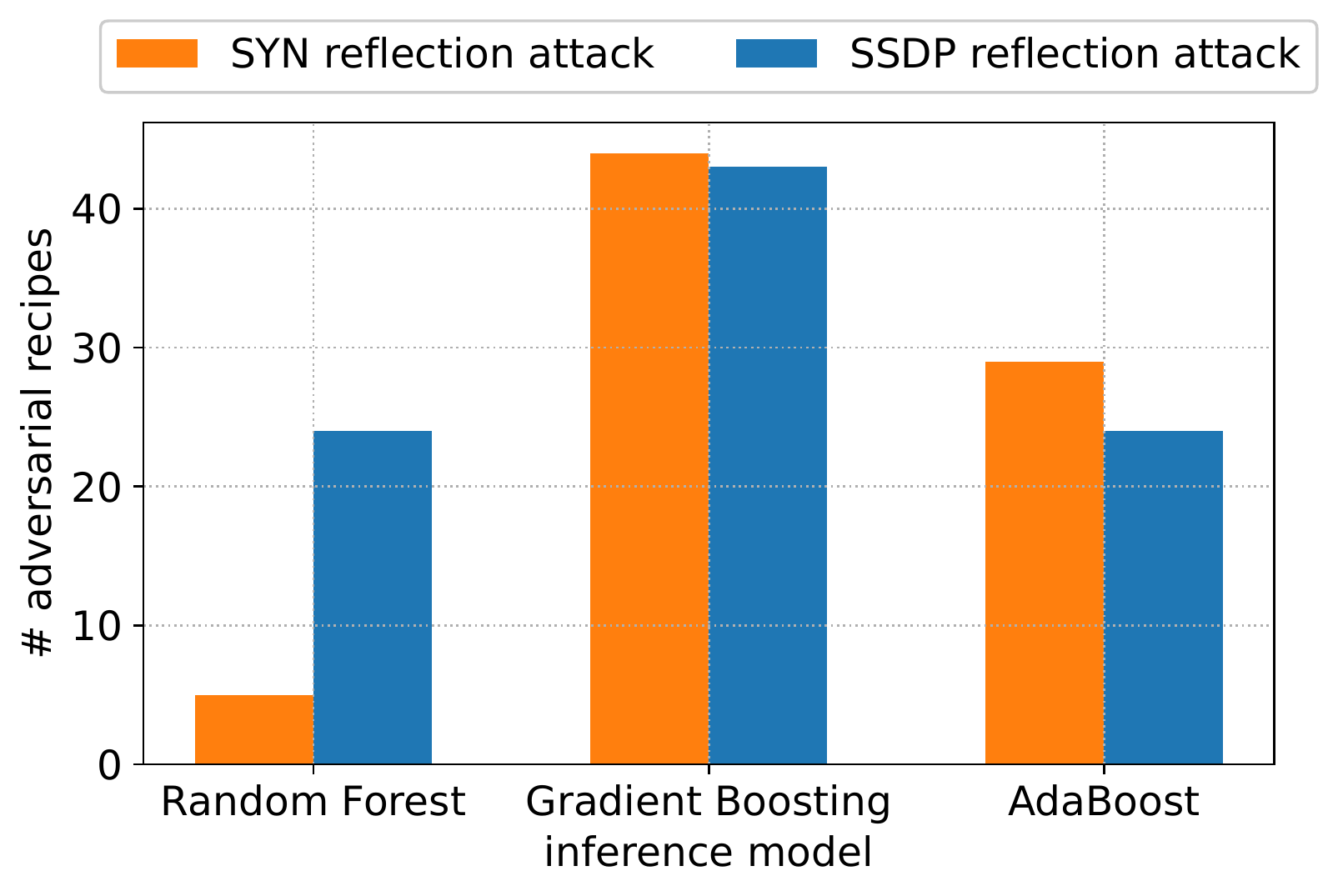}
	\vspace{-4mm}
	\caption{Number of generated adversarial recipes for three ensemble models for attack impact of 1000 packets.}
	\label{fig:recipe_count_model}
	\vspace{-1mm}
\end{figure}

\vspace{-2mm}
\subsection{The Inference Model}\label{sec:infer model}

Inspired by \cite{TNSM2020}, we consider flow-level features from network traffic, periodically computed over one-minute windows. That way, no need for inspecting packet payloads. 
Traffic features include statistics (packet and byte counts) of $\downarrow${\myverb{DNS}}, $\uparrow${\myverb{DNS}}, $\downarrow${\myverb{NTP}}, $\uparrow${\myverb{NTP}}, $\uparrow${\myverb{SSDP}}, $\downarrow${\myverb{LAN}}, $\downarrow${\myverb{WAN}}, and $\uparrow${\myverb{WAN}}; where $\uparrow$ indicates upstream traffic from each IoT device, and $\downarrow$ highlights downstream traffic to each IoT device. This means a total of 16 features.
	
We train our model with a dataset collected from our testbed (comprising nine IoT devices), during February-May 2020. Our full dataset has 954,384 benign instances. We use data from February, March and the first half of April for training, and the second half of April and May for testing.

We use the Python Scikit Learn library to train multi-class classifiers by three representative ensemble decision-tree algorithms, namely Random Forest, Gradient Boosting, and AdaBoost, to predict device class label. These models provide us with a classification score used as a baseline threshold for detecting misbehavior. 
Table~\ref{tab:modelsPerformance} shows the accuracy and false positive rate of each model.
For detecting misbehaviors, we use classification scores' $\mu$ and $\sigma$ for each class of IoT device (shown in Table~\ref{tab:conf} for Random Forest model) obtained from the training dataset in such a way that if for a given instance the model's classification score for its predicted label is below $\mu-\sigma$ (of that label), the instance is considered as malicious.
Table~\ref{tab:conf} shows the Random Forest model's classification score for each class of IoT devices in our testbed obtained from the training dataset. We use $\mu$ and $\sigma$ values obtained from the training for detecting IoT devices' misbehavior in such a way that if for a given instance the model's classification score for its predicted label is below $\mu-\sigma$ (of that label), the instance is considered as malicious.

%

\subsection{AdIoTack Offline Instances}

\begin{table}[t!]
	\centering
	\small
	\caption{Expected classification score ($\mu$ and $\sigma$) of the Random Forest model obtained from training.}
	\begin{tabular}{l|cc} \hline
		
		\hline
		& \multicolumn{2}{c}{Benign} \\ 
		IoT class & $\mu$ & $\sigma$ \\ \hline 
		Amazon Echo & 0.94 & 0.10 \\ 
		Belkin motion & 0.96 & 0.06\\ 
		Belkin switch & 0.86 & 0.18 \\ 
		Chromecast & 0.90 & 0.16 \\
		Hue bulb & 0.99 & 0.01 \\ 
		LiFX bulb & 0.72 & 0.15\\ 
		Netatmo camera & 0.88 & 0.14 \\ 
		Samsung camera & 1.00 & 0.002 \\ 
		TP-Link switch & 0.86 & 0.10\\ \hline
		\hline
	\end{tabular}	\label{tab:conf}
	\vspace{-4mm}
\end{table}

\begin{figure}[t!]
	\centering
	\includegraphics[width=0.93\linewidth,height=0.73\linewidth]{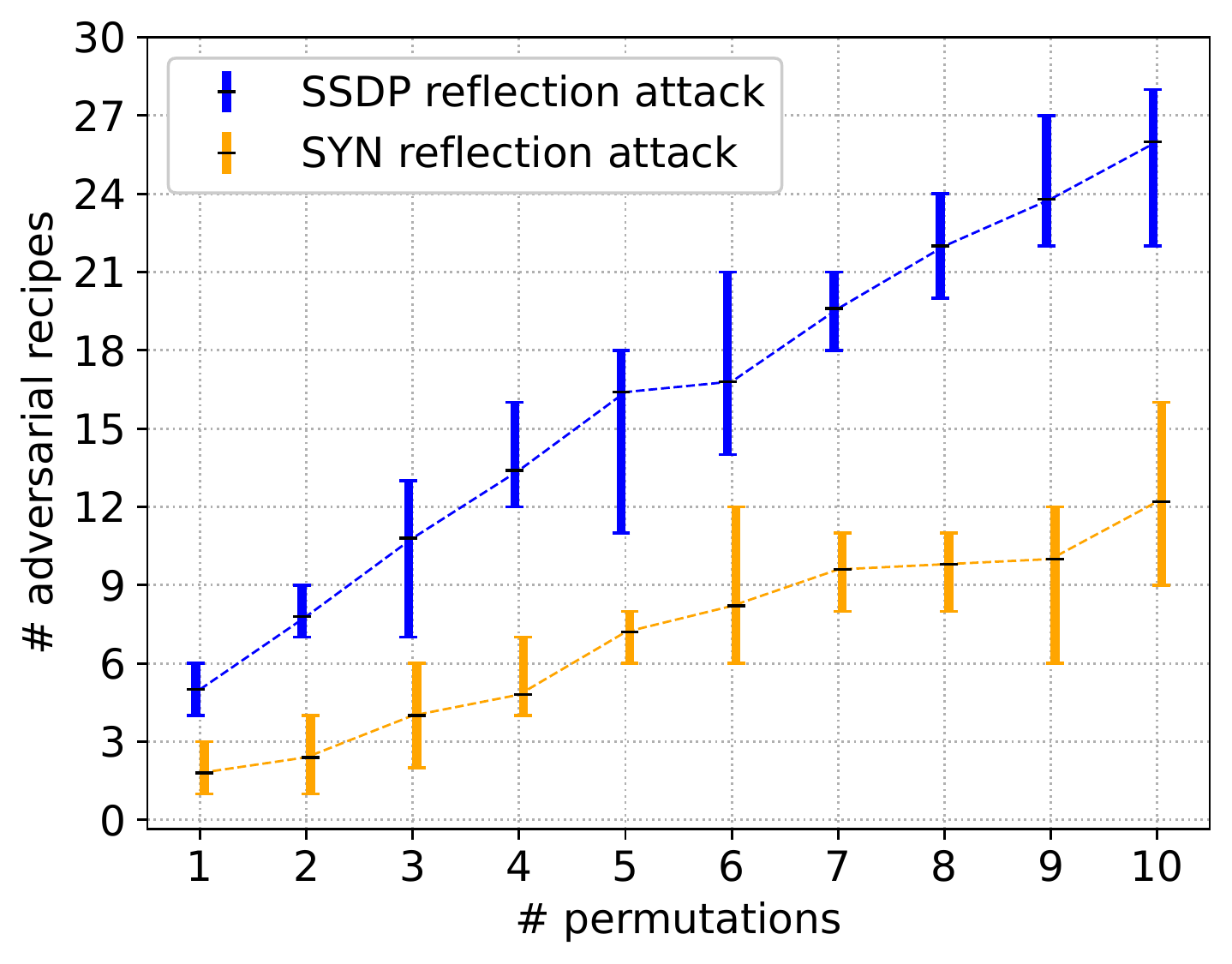}
	\vspace{-4mm}
	\caption{Number (min, max, avg) of generated adversarial recipes for different permutation counts for attack impact of 1000 packets.}
	\label{fig:recipe_permutation_count}
	\vspace{-2mm}
\end{figure}


In this part, we evaluate the performance of AdIoTack in offline adversarial learning with a view to launch two well-known network-based attacks, namely TCP SYN and SSDP reflection attacks.

Let us begin by showing that our offline learning method is generalizable by applying it to the three popular ensemble models. Fig.~\ref{fig:recipe_count_model} indicates the number of adversarial recipes for SSDP and SYN reflection attacks per inference model. Random Forest seems to be more robust to adversarial attacks compared to the other two models, corroborating with the observations in \cite{Katzir2018}. Because of its relative robustness, we choose Random Forest for the rest of the experimental evaluation in this paper.

We now quantify the impact of the sequence order of trees in generating adversarial recipes. We define permutation as a random shuffle of the trees inside the ensemble model, which we need to iterate through them in the same order as the permutation suggests. We show that various permutations can result in different adversarial recipes. Fig.~\ref{fig:recipe_permutation_count} shows the min, max, and average number of unique recipes for different permutation counts by running the algorithm five times for each permutation value. For example, if the permutation count is three, we run the offline learning with three different permutations for each device over five rounds and count the number of unique recipes at each round. The plot shows an overall increasing trend which means as we try different permutations, there is a high chance of finding new recipes. Also, the plot suggests that SSDP reflection can generate more recipes than SYN reflection, in which the increasing trend seems to become smoother after three permutations. The average time taken for generating adversarial recipes per permutation is about 1.5 minutes for SSDP reflection and 2.2 minutes for SYN reflection attack. This is probably because four features (packet and byte count of flows $\downarrow${\myverb{WAN}} and $\uparrow${\myverb{WAN}}) are affected by SYN reflection attack, whereas only two features (packet and byte count for $\uparrow${\myverb{SSDP}}) are affected by the SSDP reflection attack. 

\begin{figure}[t!]
	\centering
	\includegraphics[width=0.93\linewidth,height=0.73\linewidth]{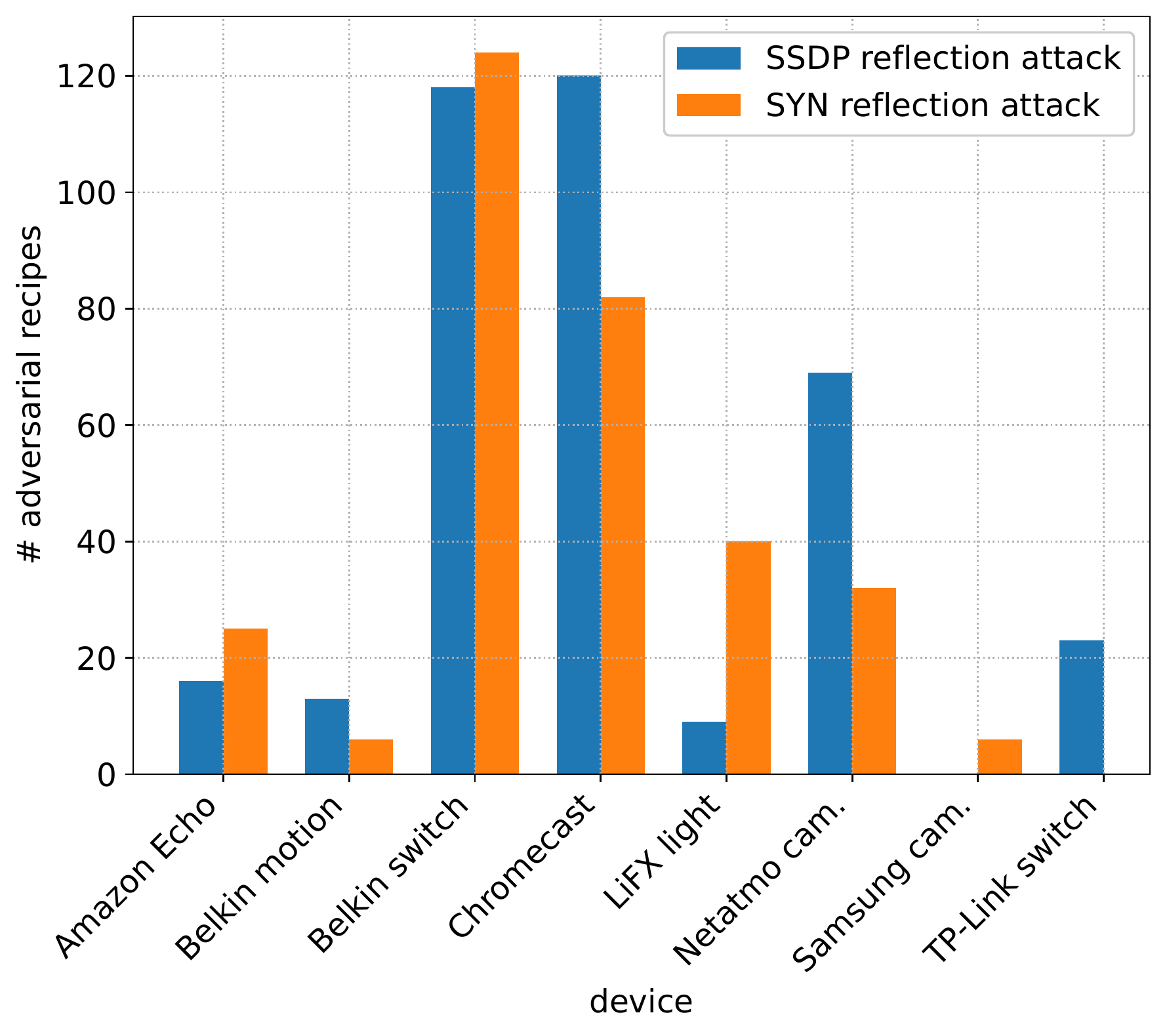}
	\\[-5mm]
	\caption{Number of generated adversarial recipes per IoT device for two representative attacks (result of a 20-permutation run).}
	\label{fig:dev_recipe_count}
	\vspace{-5mm}
\end{figure}

\begin{figure*}[t!]
	\begin{center}
		\mbox{
			\subfigure[Amazon Echo.]{
				{\includegraphics[width=0.475\textwidth]{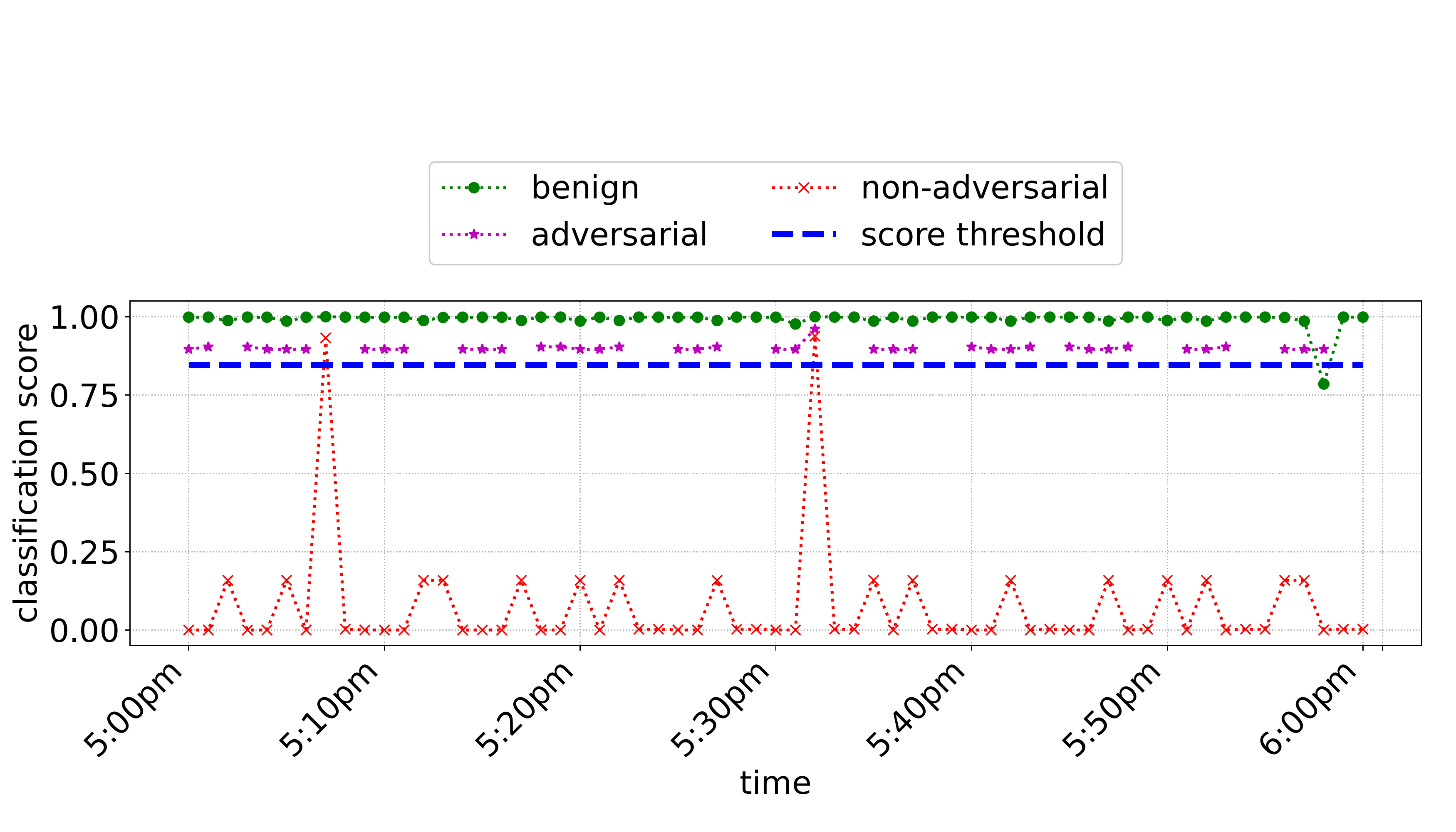}}\quad
			}
		}
		\hspace{-8mm}
		\mbox{
			\subfigure[Belkin switch.]{
				{\includegraphics[width=0.475\textwidth]{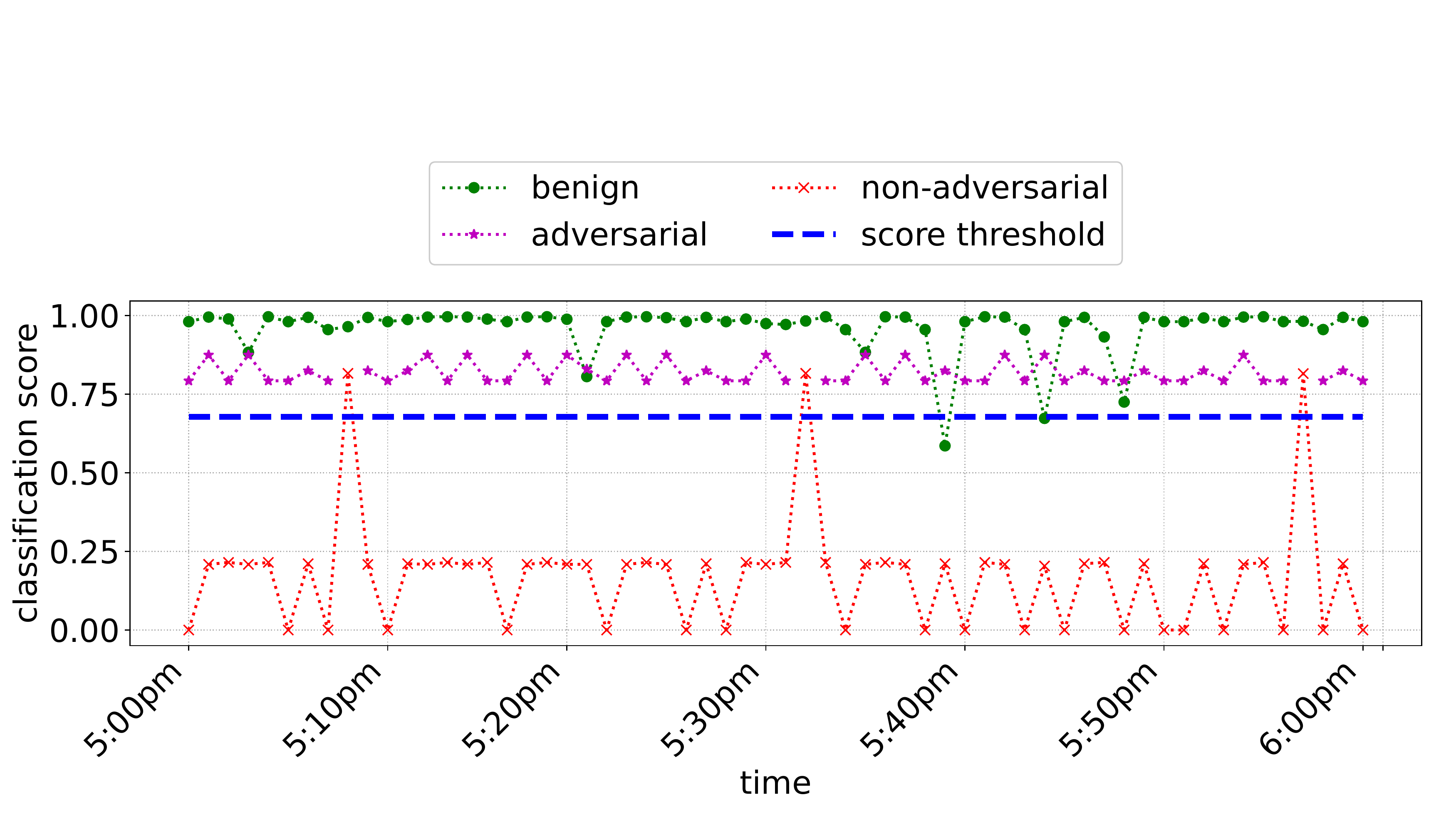}}\quad
			}
		}
		\vspace{-5mm}
		\caption{Time trace of model performance against replayed instances of two representative IoT devices: (a) Amazon Echo, and (b) Belkin switch, in three scenarios: benign (green), synthetic adversarial SYN reflection attack (purple) and synthetic non-adversarial SYN reflection attack (red).}
		\label{fig:replay_attack}
		\vspace{-4mm}
	\end{center}
\end{figure*}


Fig.~\ref{fig:dev_recipe_count} illustrates the number of generated recipes for each IoT device in our testbed by running the offline learning algorithm with 20 permutations. Belkin switch and Chromecast in both types the attacks. Samsung smart camera has no recipe for SSDP reflection, TP-Link has no recipe for SYN reflection attack, and Hue lightbulb has no recipe for both of the attacks, indicating tight constraints for those classes.

Before experimenting with our adversarial recipes in an operational network of IoT devices, we evaluate the performance of our inference model and efficacy of offline adversarial instances by replaying various unseen datasets in three scenarios, namely: (1) a purely benign dataset is expected to receive high classification scores from the model, (2) syntactically changed features of the benign instances (to represent non-adversarial SYN reflection attack) are expected to receive low scores from the model (highlighting a detection), and (3) synthetically changed features of the benign instances (to represent adversarial SYN reflection attack) are expected to receive high scores from the model (highlighting a miss).


Fig.~\ref{fig:replay_attack} shows the results for two representative IoT devices over an hour. As expected, almost all benign instances are classified with a high score (\ie above the threshold shown by dashed blue line), highlighting the expected performance under normal situations (benign traffic). The dotted red curve shows the model's score for non-adversarial attack instances, which the model detects a majority of them (receiving scores below the threshold). Finally, the purple curve highlights the model's inability to detect adversarial attacks (\ie giving scores above the threshold, meaning benign). In some epochs like at 5:03pm for Amazon Echo and 5:08pm for Belkin switch, there is no recipe in the feasible region to project the current instance, thus missing data points.

\begin{figure}[t!]
	\centering
	\includegraphics[width=0.95\linewidth]{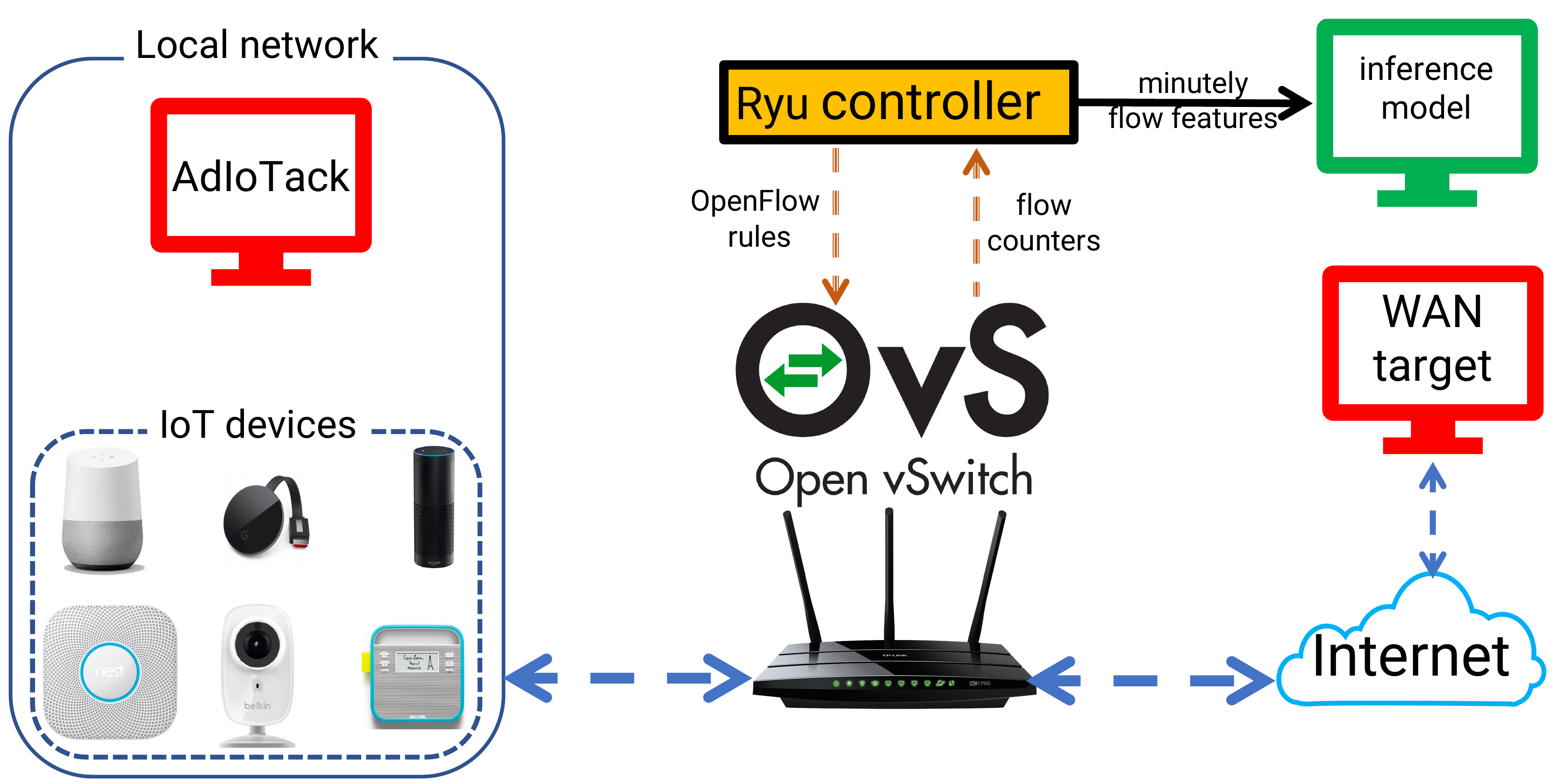}
	\\[-2mm]
	\caption{Prototype of AdIoTack.}
	\label{fig:setup}
	\vspace{-4mm}
\end{figure}

\subsection{AdIoTack Online Evaluation}

To demonstrate adversarial attacks in a realistic scenario, we deploy our inference model in a system that can classify traffic in real-time. It receives a stream of flow-based telemetry and predicts traffic features of individual connected devices every minute. To collect real-time telemetry, we use an SDN-enabled home gateway in our setup, shown in Fig.~\ref{fig:setup}. We installed Open Virtual Switch (OVS) on the gateway of our IoT network. Ryu, an open-source SDN controller, is installed on another machine that communicates with this OVS. The Ryu controller initializes the setup by sending a fixed set of flow rules specific to each IoT device on the network, of which the inference model needs the corresponding flow counters as features (features described in \S \ref{sec:infer model}). Every minute, the model receives flow counters from the controller and classifies individual devices. The model also detects unexpected behavior by way of classification score (explained in \S \ref{sec:infer model}).

\begin{figure*}[htb]
	\begin{center}
		\mbox{
			\subfigure[Belkin switch.]{
				{\includegraphics[width=0.43\textwidth]{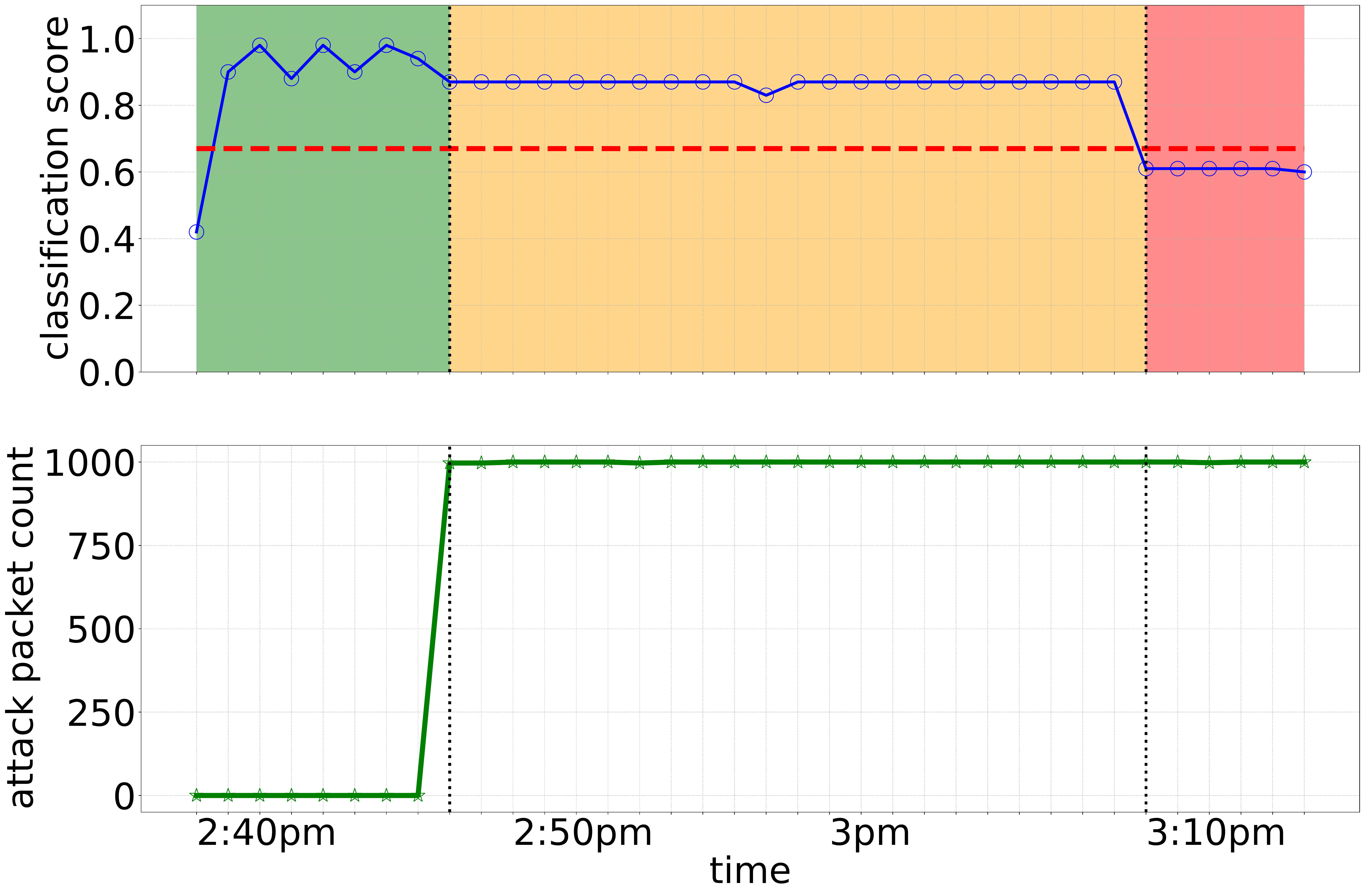}}\quad
			}
		}
		\hspace{-8mm}
		\mbox{
			\subfigure[Chromecast.]{
				{\includegraphics[width=0.43\textwidth]{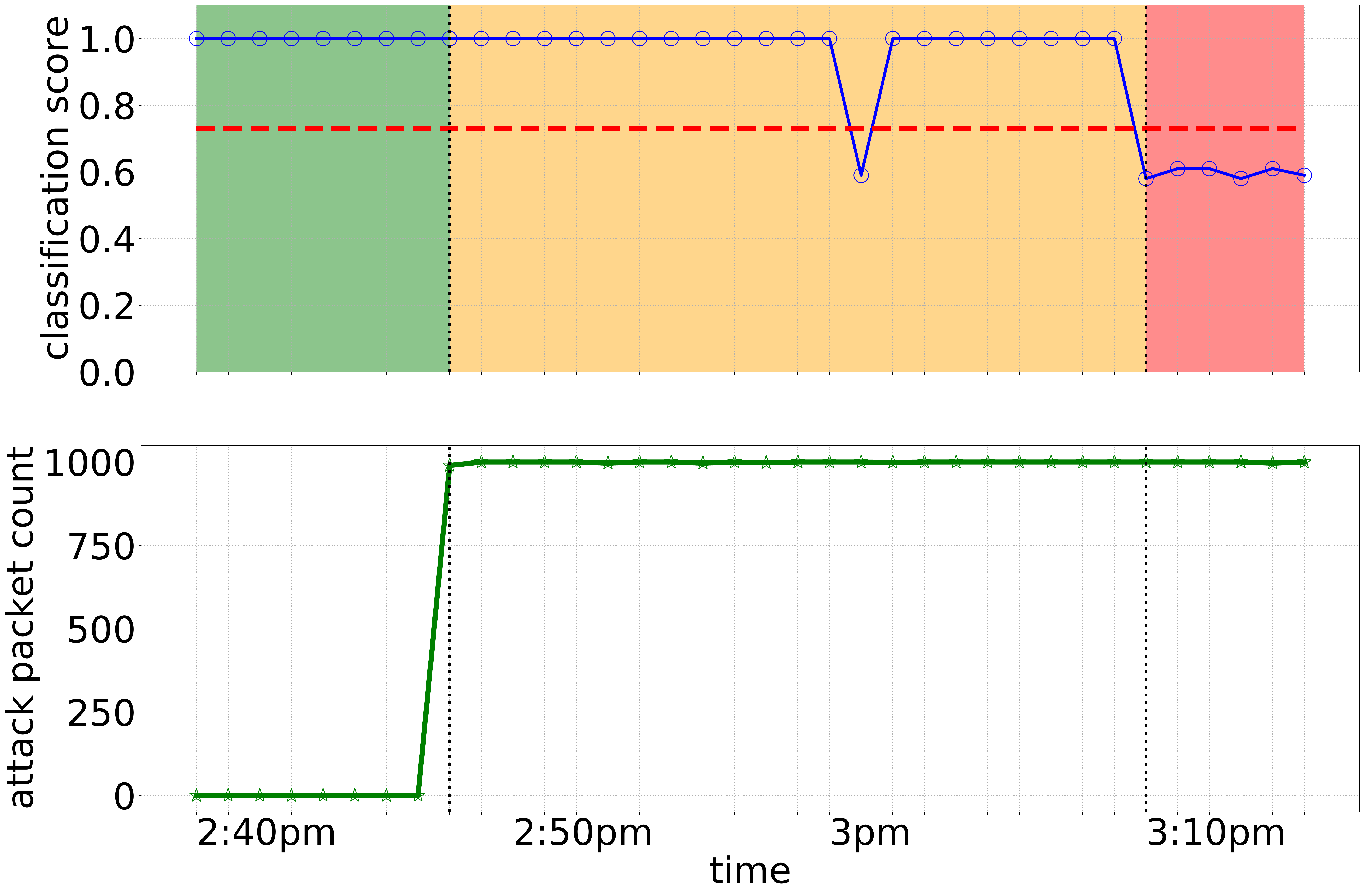}}\label{fig:ExecChrom}\quad
			}
		}
		\mbox{
			\subfigure[Netatmo camera.]{
				{\includegraphics[width=0.43\textwidth]{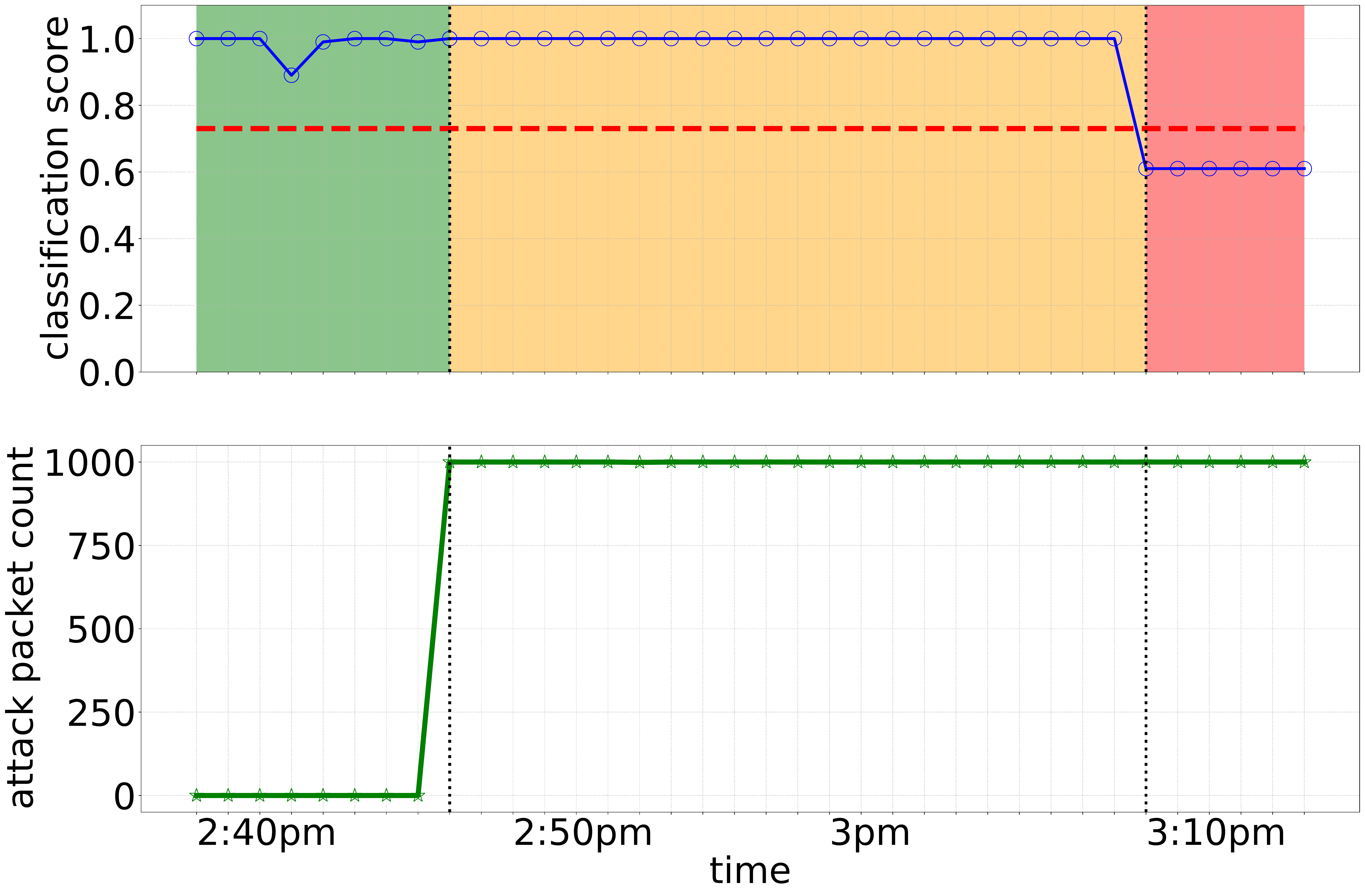}}\quad
			}
		}
		\hspace{-8mm}
		\vspace{-2mm}
		\caption{Time trace of model performance and attack traffic on three IoT devices in three scenarios: normal operation (green), adversarial attack (orange) and non-adversarial attack (red).}
		\label{fig:time_trace}
		\vspace{-7mm}
	\end{center}
\end{figure*}

Fig.~\ref{fig:time_trace} shows the performance of AdIoTack in the online execution phase for three representative devices with SYN reflection attack. 
Our entire evaluation lasted for 35 minutes with three stages: between 2:40pm and 2:48pm (highlighted in green), no attack was executed to observe the behavior of our inference model in normal operation; between 2:48pm and 3:10pm (highlighted in orange), we performed adversarial SYN reflection attacks in each epoch; lastly, between 3:10pm and 3:15pm (highlighted in red), we performed non-adversarial SYN reflection attack with the same impact of the adversarial ones launched in the middle stage. The upper subplots show the inference model's classification score at the end of each epoch. The score threshold of each class is shown by a constant dashed red line. The lower subplots show the number of attack packets received by the ultimate target machine (victim server) residing on the WAN interface of the gateway (outside the local IoT network).

\begin{figure*}[htb]
	\begin{center}
		\mbox{
			\subfigure[Belkin motion sensor.]{
				{\includegraphics[width=0.435\textwidth]{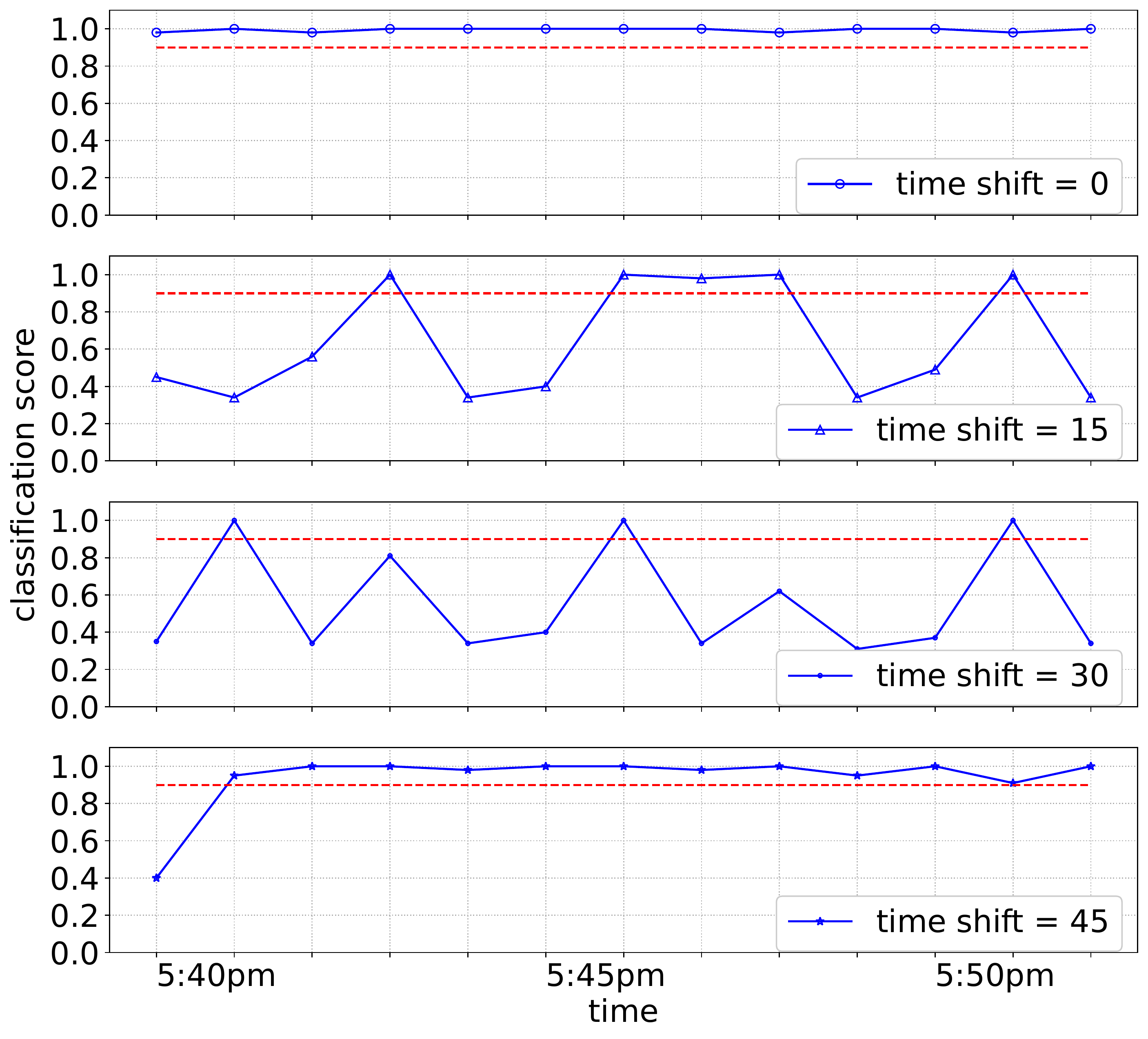}}\quad
			}
		}
		\mbox{
			\subfigure[Chromecast.]{
				{\includegraphics[width=0.435\textwidth]{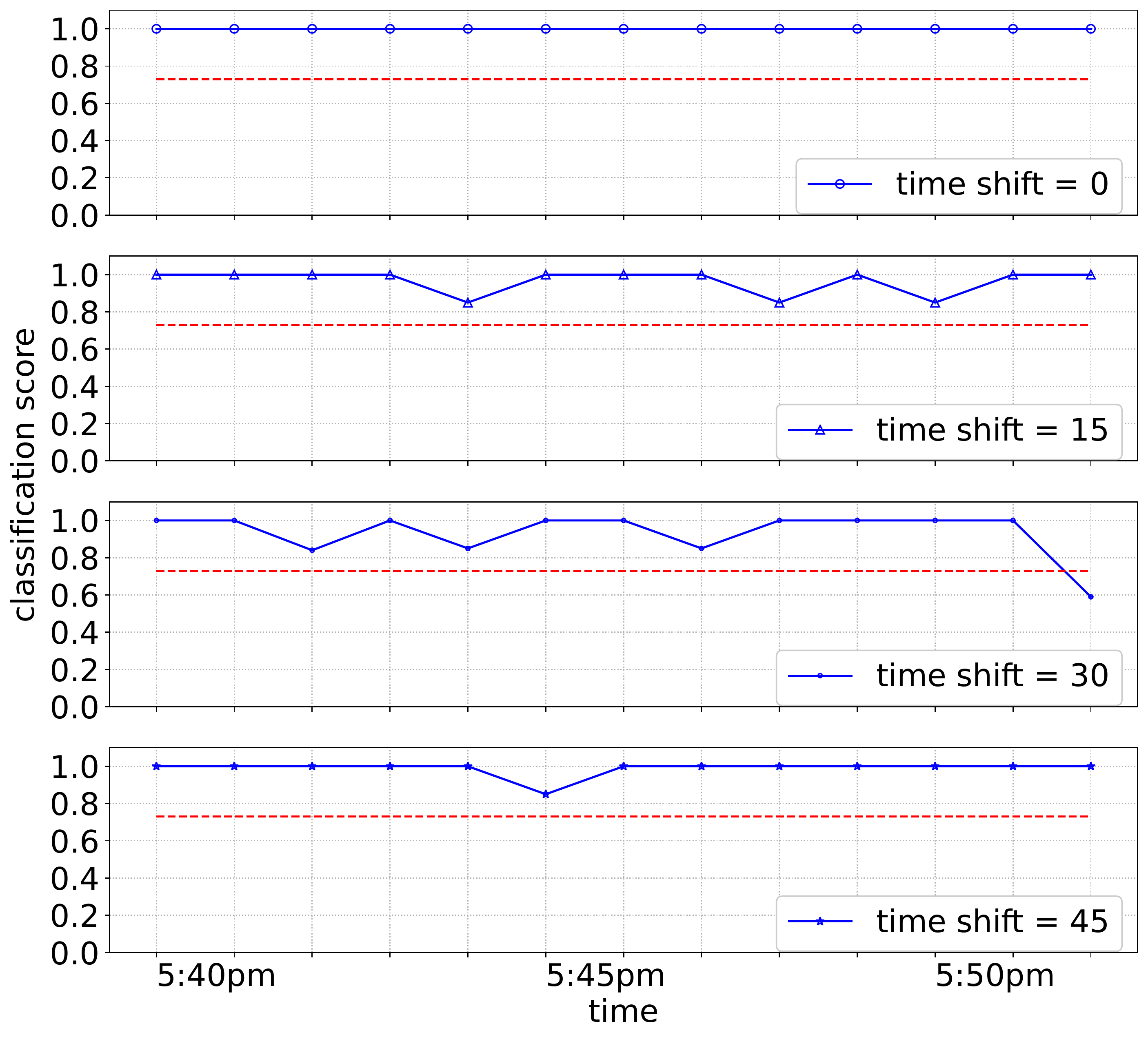}}\quad
			}
		}
		\vspace{-3mm}
		\caption{Effect of time-shift for two IoT devices.}
		\label{fig:time_shift}
		\vspace{-8mm}
	\end{center}
\end{figure*}

It can be seen that the inference model displays an acceptable performance under normal operation (\ie pure benign traffic, without any attack). Moving to the orange regions, we observe that all adversarial reflection attacks are successful (classification scores above the threshold), except during an epoch for Chromecast in Fig.~\ref{fig:ExecChrom}.
Lastly, focusing on the red regions, the model gives scores lower than expected threshold to non-adversarial instances across the three representative devices.


During the online evaluation, we found that it is impossible to execute any adversarial attack on three devices, namely Samsung smart camera, Philips Hue lightbulb, and Amazon Echo. For Samsung smart camera, the reason is that no feasible adversarial recipe was found for $I_c$ (device's current feature vector) to be projected on as $I_c$ during our experiment violated the upper bound of the rule on $\uparrow${\myverb{SSDP}} feature. For the other two devices, the reason relies on the network aspect. Philips Hue lightbulb responds to the corrective ping message sourced by AdIoTack. However, the response does not revert the gateway's MAC table; probably because Philips Hue is not a standalone device and relies on a separate bridge for communication. Amazon Echo did not respond to the ping message, hence the disrupted MAC table issue stopped us from launching the attack.

During this evaluation, we assumed that AdIoTack knows the inference cycles of the model \ie when the model fetches the flow counters from the SDN controller and classifies the traffic. This is a valid assumption as we already declared our work to be a white-box approach. Still, we further evaluate AdIoTack online execution in scenarios where inference cycles are unknown. Fig.~\ref{fig:time_shift} illustrates this evaluation for Belkin motion sensor and Chromecast. We launched adversarial attacks on these devices with four different timings. Time-shift of $T$ means that the attacker is $T$ seconds ahead of the inference model. We need to note that the time window in this evaluation is 60 seconds meaning that the model becomes online every 60 seconds.

Unsurprisingly, Fig.~\ref{fig:time_shift} shows that adversarial attacks with the time shift of zero are $100$\% successful in bypassing the model. Fig.~\ref{fig:time_shift} shows that for Chromecast, time-shift almost has no effect on the success of adversarial attacks; however, for Belkin motion sensor, it has significant effects. The results suggest that a time shift of 30 seconds leads to the minimum success rate ($23$\%), but when it gets closer to the inference model's time cycle, \ie time shift of 45 seconds, the success rate significantly rises ($92$\%). The key takeaway is that regardless of when AdIoTack launches the attack, there is always a chance to bypass the model.

\section{Refining Resilience of Decision Trees Against Adversarial Volumetric Attacks}\label{sec:patching}

We have so far highlighted the vulnerability of decision tree-based models against systematically crafted adversarial attacks. In this section, we develop a method called \textit{patching} to refine the model post-training, making decision trees robust against adversarial volumetric attacks.

\begin{figure}[t!]
	\centering
	\includegraphics[width=0.95\linewidth]{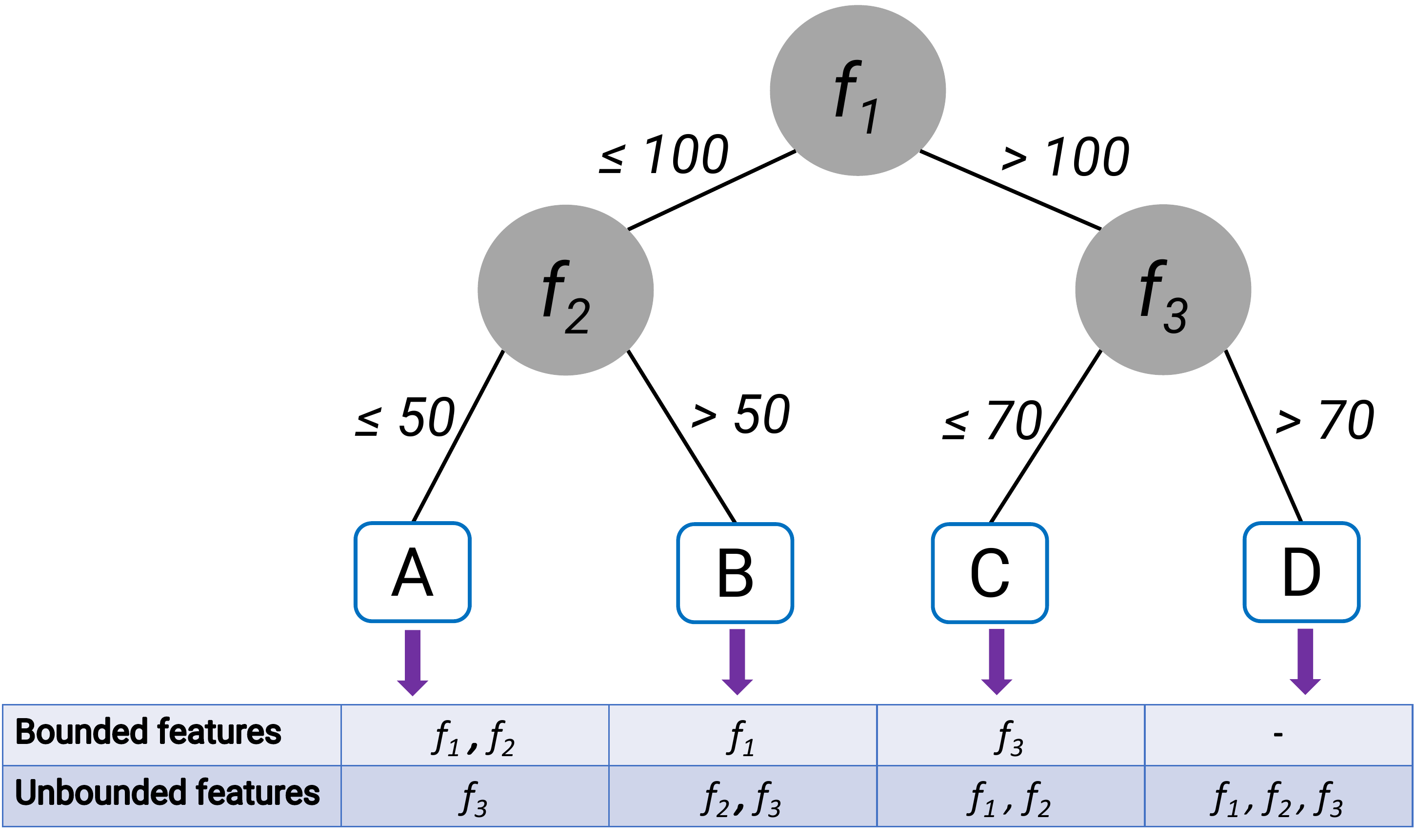}
	\\[-2mm]
	\caption{Bounded and unbounded features for each leaf in a decision tree.}
	\label{fig:vulnerable_tree}
\end{figure}

Decision trees trained purely by benign traffic instances are inherently vulnerable (given their structure) to volumetric attacks. Fig.~\ref{fig:vulnerable_tree} illustrates an illustrative decision tree with four leaves and three decision nodes on three features. For each leaf, we define \textit{bounded} and \textit{unbounded} feature sets. For a given leaf $l$, its bounded feature set includes feature $f$, if inside the path from the tree's root to $l$, there is at least one decision node on $f$ which must take the left branch (\ie $f \le \tau$) to reach $l$. Otherwise, if there is no such a decision node for $f$, we consider $f$ as an unbounded feature for $l$. For example, in Fig.~\ref{fig:vulnerable_tree}, consider the leaf with label `{\myverb{C}}'. To reach this leaf from the root, the following conditions must be met: $f_1>100$ and $f_3\le70$. Thus if an instance has $f_1=+\infty$, $f_2=+\infty$, and $f_3=70$, it still reaches the leaf with no issue as there is no upper boundary check for $f_1$ and there is no upper or lower boundary check for $f_2$ at all.

In the literature, there exist methods \cite{chen2019robust, calzavara2020treant} for developing robust decision trees against adversarial attacks. The current techniques manipulate the training process of decision trees by changing the original model to make sure if an adversarial attack happens, the model still gives the correct output label which is desirable in image classification which requires a given set of adversarial instances to be provided before refining the model \ie the model still might be vulnerable to another set of adversarial instances. That way, they sacrifice the prediction accuracy for benign instances, in order to make decision trees robust in adversarial scenarios. Also, given their primary focus is on image recognition, they do not have a notion of volumetric attacks, and thus, unbounded features remain loose. Our method, however, focuses on a different problem (\ie volumetric cyberattacks), and aims to refine a trained model. Therefore, it does not interfere with the training process and does not require re-training. Our method receives a trained decision tree-based model and then patches the unbounded features with no dependency on any adversarial recipes. If the training dataset gives a correct representation of devices' traffic characteristics by at least capturing the maximum traffic volume for each class correctly, the patched model will have the exact same accuracy/false positive over benign instances as the original model. However, if there are data instances in the testing dataset that have higher traffic features' value than the limits which were captured from the training dataset, our method would flag them an as malicious (\ie higher false positives than the original unpatched model).

\begin{figure}[t!]
	\centering
	\includegraphics[width=0.95\linewidth]{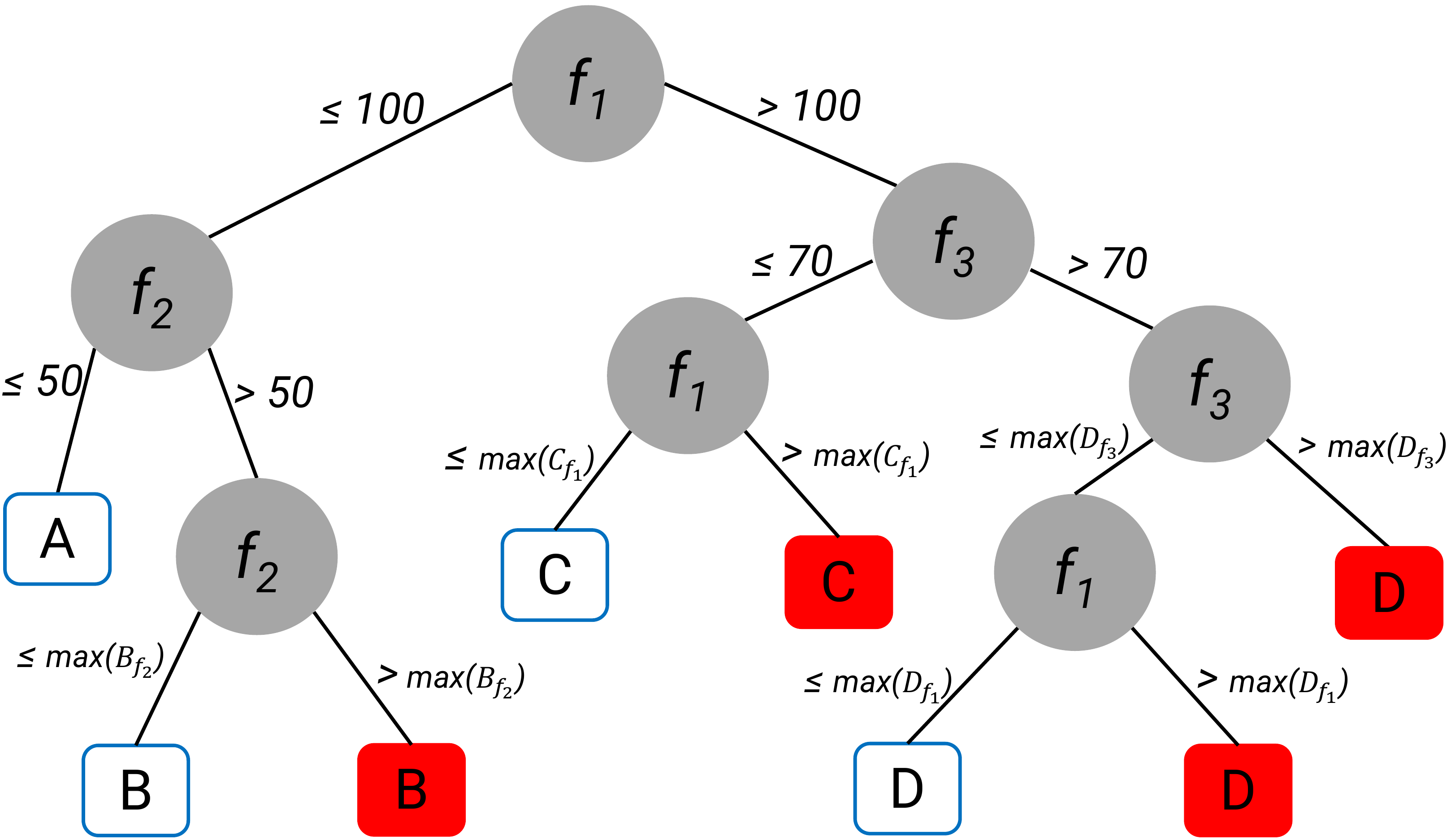}
	\caption{Essential patching performed on individual leaves of the vulnerable decision tree from Fig.~\ref{fig:vulnerable_tree}.}
	\label{fig:patched_tree}
\end{figure}

We define a patch for feature $f$ on leaf $l$ as follows:
\vspace{-1mm}
\begin{gather*}
\label{eq:patch}
patch_{l, f} = f \le max(l_f)
\end{gather*}

This patch can be seen as a new decision node added just before reaching the leaf, that checks the upper bound of $f$ for $l$. The upper bound is calculated based on the training dataset used for training the original model. As an example, based on our training dataset, Amazon Echo sends a maximum of four DNS packets every minute; thus, $max(Amazon~Echo_{\uparrow DNS~pkt}) = 4$. Any instance $x$ reaching leaf $l$, which $x_f>max(l_f)$, goes to a new leaf that yields $x$ as a malicious instance.

There are two noteworthy points: (1) Because IoT devices send/receive a low volume of traffic and have repetitive behavioral patterns, individual devices' maximum traffic volume can be captured \cite{Sivanathan2017}. However, this could be challenging for personal computers and smartphones which their traffic volume highly depends on users' activities. (2) Our method reduces the impact of adversarial volumetric attacks significantly. However, low-impact attacks (\ie within the normal behavior range of IoT devices) can still bypass the model.

\begin{figure}[t!]
	\centering
	\includegraphics[width=0.95\linewidth,height=0.75\linewidth]{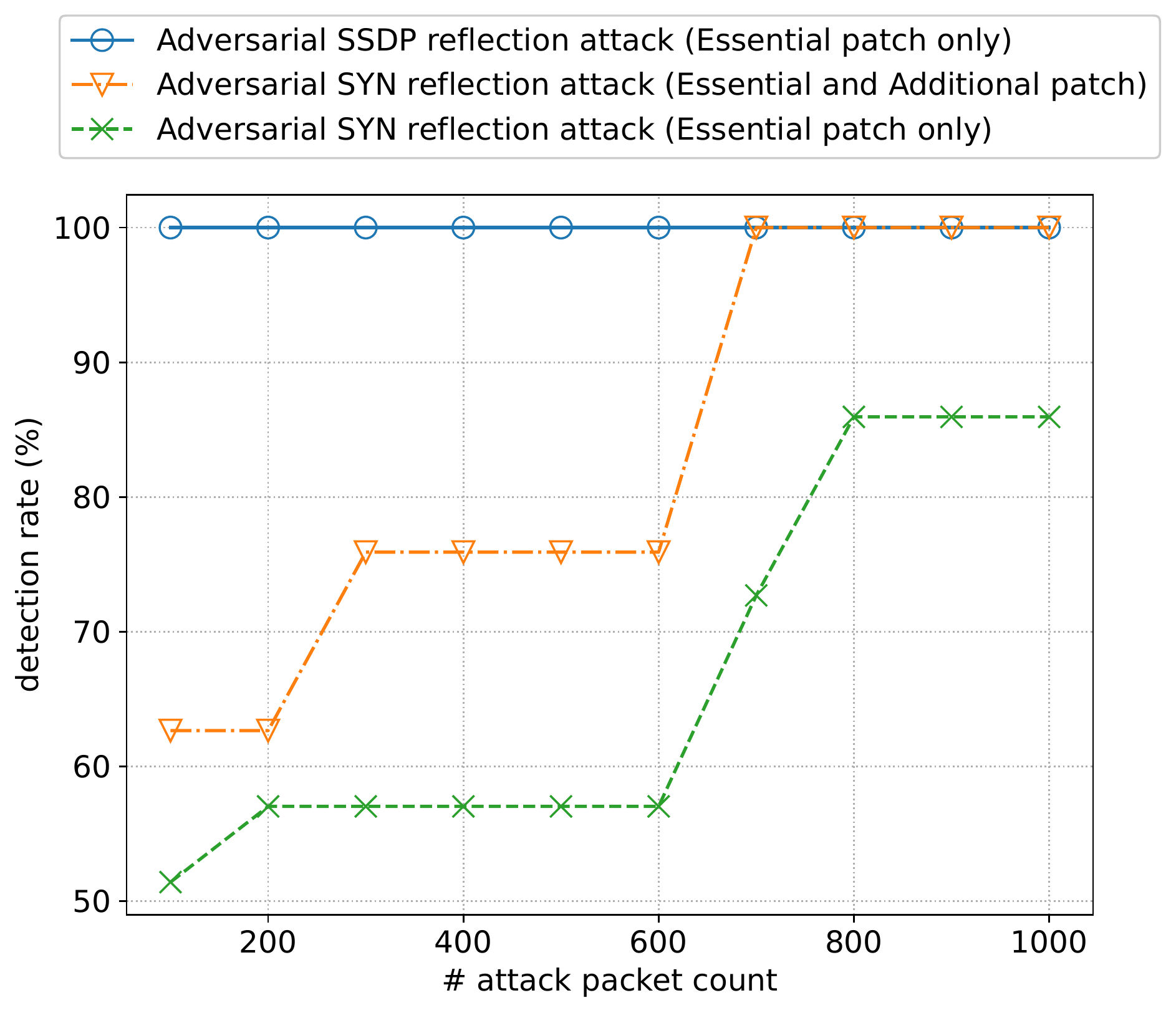}
	\vspace{-5mm}
	\caption{Detection rate of Essential and Additional patching for adversarial SYN and SSDP reflection attacks.}
	\label{fig:detection_rate}
\end{figure}

There are two scenarios where feature $f$ can become unbounded for leaf $l$: (1) When there is a decision node on $f$, but its right branch (\ie $f > \tau$) is taken on the path from the tree's root to $l$ (\eg $f_1$ for leaf `{\myverb{C}}' in Fig.~\ref{fig:vulnerable_tree}); and (2) When there is no decision node on $f$ through the path at all (\eg $f_2$ for leaf `{\myverb{C}}' in Fig.~\ref{fig:vulnerable_tree}). We develop two patching techniques to address these two scenarios. \textit{Essential Patching} solves the first problem by traversing a given vulnerable decision tree and patching leaves with unbounded features through paths from the root. Fig.~\ref{fig:patched_tree} shows the result of essential patching performed on the tree from Fig.~\ref{fig:vulnerable_tree}. Adversarial instances will be routed to the red leaves, which results in true detection of the attack.

Although essential patching is necessary to fix some unbounded features, it does not cover the second scenario where some features do not even exist on some paths. For example, leaf ``{\myverb{A}}'' in Fig.~\ref{fig:patched_tree}, still can be targeted with volumetric attack on $f_3 = +\infty$. In fact, some adversarial attacks can be captured solely with essential patching (\eg SSDP reflection attack in Fig.~\ref{fig:detection_rate}). To overcome this problem, we define \textit{Additional Patching} which patches all leaves for a given feature. This way, regardless of the trees' structure, we make sure all leaves are patched against adversarial volumetric attacks on the desired features. Additional patching can be done in different ways; for example, one may choose to patch all leaves for all possible features to create maximum robustness,  making the model complex and slow. Another one may use expert knowledge to select specific vulnerable features to limit the complexity of the model while making it robust against attacks over those features that are more likely to be targeted by attackers.
After applying the essential patching to our original model, we quantify the accuracy and false-positive over the testing dataset and they are 96\% and 0.09\%, respectively, which are exactly the same as the original model.

Fig.~\ref{fig:detection_rate} shows the performance of our patching methods over adversarial SSDP and SYN reflection attacks with a range of attack impacts (100 to 1,000 packet count). Note that these attack instances are created based on adversarial recipes, bypassing the unpatched model, as shown in the previous section. Essential patching is sufficient to detect all SSDP reflection attacks while detecting $86$\% of the high-impact SYN reflection attacks. Having an additional patch over $\uparrow$ {\myverb{WAN}} packet count feature increases the detection rate for low, medium, and high impact attacks; and detects all attacks with an impact greater than $636$ packets. The average impact of SYN reflection attacks that can still bypass the model is $117$ packets across our IoT devices, with the maximum impact of $636$ packets for Chromecast. Compared to the vulnerable model, which has no limit over the attack packet count (\ie unlimited impact), this impact shows a significant improvement in terms of robustness.

\begin{figure}[t!]
	\centering
	\includegraphics[width=0.99\linewidth]{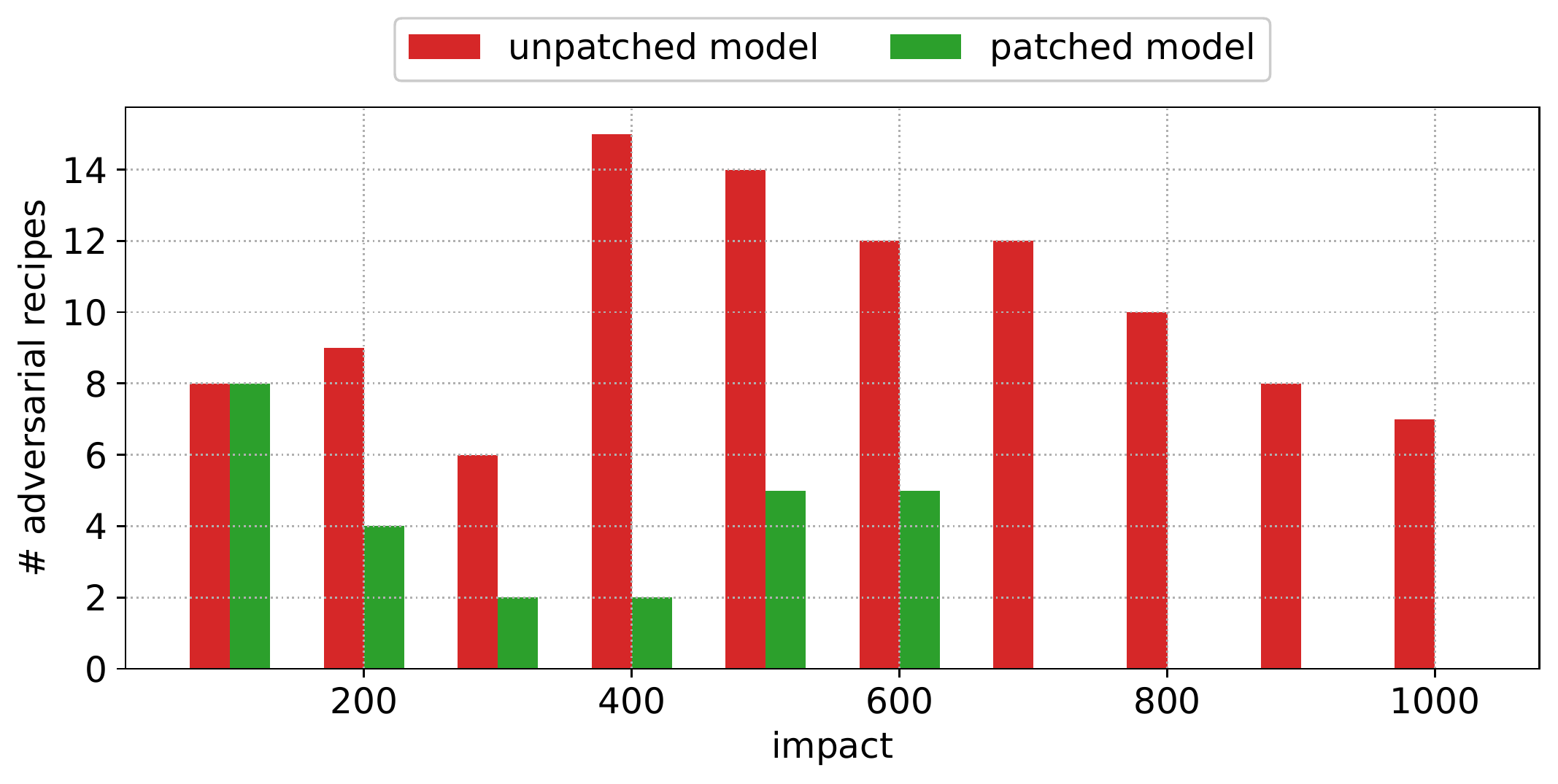}
	\vspace{-8mm}
	\caption{Number of recipes generated for SYN reflection attack before and after patching the model.}
	\label{fig:patched_vs_unpatched}
	\vspace{-3mm}
\end{figure}

Finally, we validate the efficacy of our patched model by re-running the adversarial offline learning function on the refined model. Fig.~\ref{fig:patched_vs_unpatched} compares the number of generated recipes in the patched model versus the original (unpatched) model. The patched model (green bars) has become more robust against SSDP reflection attacks with any impact. The improvement becomes more evident as the impact increases. More importantly, no recipe exists for volumetric attacks with an impact above 600 packets.


\section{Conclusion}

In this paper, we developed \textit{AdIoTack}, a systematic way of quantifying and refining the resilience of decision tree ensemble models against data-driven adversarial attacks.
We first developed a white-box algorithm that automatically generates recipes of volumetric network-based attacks that can bypass the inference model unnoticed. Our algorithm takes the intended attack on a victim class and the model as inputs. We developed a systematic method to successfully launch the intended attack on real networks. We next prototyped AdIoTack and validated its efficacy on a real testbed of real IoT devices monitored by a trained Random-Forest model. We demonstrated how the model detects all non-adversarial volumetric attacks on IoT devices while missing many adversarial ones. Finally, we developed systematic methods to patch loopholes in trained decision tree ensemble models. We demonstrated how our refined model detects 92\% of adversarial volumetric attacks.

\bibliographystyle{IEEEtranS}
\balance
\bibliography{AdIoTack}

\end{document}